\documentclass[journal]{IEEEtran}

\usepackage{float}
\usepackage{cuted} 
\usepackage{stfloats}

\usepackage{blindtext}
\usepackage{amsmath, amssymb, amsthm} 
\usepackage[dvipsnames]{xcolor}
\usepackage{hyperref}

\usepackage{color, soul} 
\usepackage{comment}
\usepackage{todonotes}

\usepackage{wrapfig}
\usepackage{siunitx}
\usepackage{bm}

\usepackage[utf8]{inputenc} 
\usepackage[T1]{fontenc}    

\usepackage{hyperref}       
\usepackage{url}            
\usepackage{booktabs}       
\usepackage{amsfonts}       
\usepackage{amssymb,amsmath}
\usepackage{nicefrac}       
\usepackage{microtype}      
\usepackage{makeidx}
\usepackage{graphicx} 
\usepackage[]{algorithm2e}

\usepackage{makecell} 
\usepackage{multirow}
\usepackage{flushend}
\usepackage{cite}
\usepackage{bbm}

\definecolor{cadmiumgreen}{rgb}{0.0, 0.42, 0.24} 

\usepackage{tikz} 
\usepackage{pifont}
\newcommand{\cmark}{{\color{cadmiumgreen}\ding{51}}}%
\newcommand{\xmark}{{\color{red}\ding{55}}}%

\newcolumntype{?}{!{\vrule width 1pt}} 
\DeclareMathOperator*{\argmax}{arg\,max}
\DeclareMathOperator\erf{erf}

\begin{document}

\title{
Towards a Comprehensive Theory \\ of Reservoir Computing 
}

\author{Denis Kleyko, Christopher J. Kymn, E. Paxon Frady,  Amy Loutfi, and Friedrich T. Sommer\\ 
\thanks{DK acknowledges funding from the European Union's Horizon 2020 research and innovation programme under the Marie Skłodowska-Curie Actions (Grant No. 839179), the Swedish Strategic Research Foundation under the Future Research Leaders program (Grant No. FFL24-0111), and the Swedish Research Council under the Starting Grant program (Grant No. 2025-05421).
The work of AL and DK was supported by Knut and Alice Wallenberg Foundation under the Wallenberg Scholars program (Grant No. KAW2023.0327).
The work of CJK was supported by the Center for the Co-Design of Cognitive Systems (CoCoSys), one of seven centers in JUMP 2.0, a Semiconductor Research Corporation (SRC) program sponsored by DARPA, in addition to the NDSEG Fellowship, Fernström Fellowship, Swartz Foundation, and NSF Grants 2147640 and 2313149.
FTS was supported by NSF Grant IIS1718991, NIH Grant R01-EB026955, and by the Kavli Foundation. 
This work was supported in part by the AFOSR under award number FA8655-25-1-7007.
}
\thanks{D. Kleyko is with the AI, Robotics and Cybersecurity Center (ARC), and with the Department of Computer Science at Örebro University, 70281 Örebro, Sweden and also with Intelligent Systems Lab at RISE Research Institutes of Sweden, 16440 Kista, Sweden (\mbox{e-mail}: \mbox{denis.kleyko@oru.se}).}%
\thanks{C. J. Kymn is with the Redwood Center for Theoretical Neuroscience at the University of California at Berkeley, CA 94720, USA (\mbox{e-mail}: \mbox{cjkymn@berkeley.edu}). }
\thanks{E. P. Frady is with the Neuromorphic Computing Lab, Intel Labs, Santa Clara, CA 95054, USA. \mbox{e-mail}: \mbox{e.paxon.frady@intel.com}}
\thanks{A. Loutfi is with the AI, Robotics and Cybersecurity Center (ARC), and with the Department of Computer Science at Örebro University, 70281 Örebro, Sweden and also with Department of Science and Technology at Linköping University, 58183 Linköping, Sweden (\mbox{e-mail}: \mbox{amy.loutfi@oru.se}).}
\thanks{F. T. Sommer is with the Redwood Center for Theoretical Neuroscience at the University of California at Berkeley, CA 94720, USA  and also with the Neuromorphic Computing Lab, Intel Labs, Santa Clara, CA 95054, USA (\mbox{e-mail}: \mbox{fsommer@berkeley.edu}). }
}

\maketitle

\begin{abstract}

In reservoir computing, an input sequence is processed by a recurrent neural network, the reservoir, which transforms it into a spatial pattern that a shallow readout network can then exploit for tasks such as memorization and time-series prediction or classification. Echo state networks (ESN) are a model class in which the reservoir is a traditional artificial neural network. This class contains many model types, each with sets of  hyperparameters. 
Selecting models and parameter settings for particular applications requires a theory for predicting and comparing performances. 
Here, we demonstrate that recent developments of perceptron theory can be used to predict the memory capacity and accuracy of a wide variety of ESN models, including reservoirs with linear neurons, sigmoid nonlinear neurons, different types of recurrent matrices, and different types of readout networks. Across thirty variants of ESNs, we show that empirical results consistently confirm the theory's predictions.
As a practical demonstration, the theory is used to optimize memory capacity of an ESN in the entire joint parameter space.
Further, guided by the theory, we propose a novel ESN model with a readout network that does not require training, and which outperforms earlier ESN models without training. 
Finally, we characterize the geometry of the readout networks in ESNs, which reveals that many ESN models exhibit a similar regular simplex geometry as has been observed in the output weights of deep neural networks.

\end{abstract}


\begin{figure}[tb]
\centering
\includegraphics[width=0.75\columnwidth]{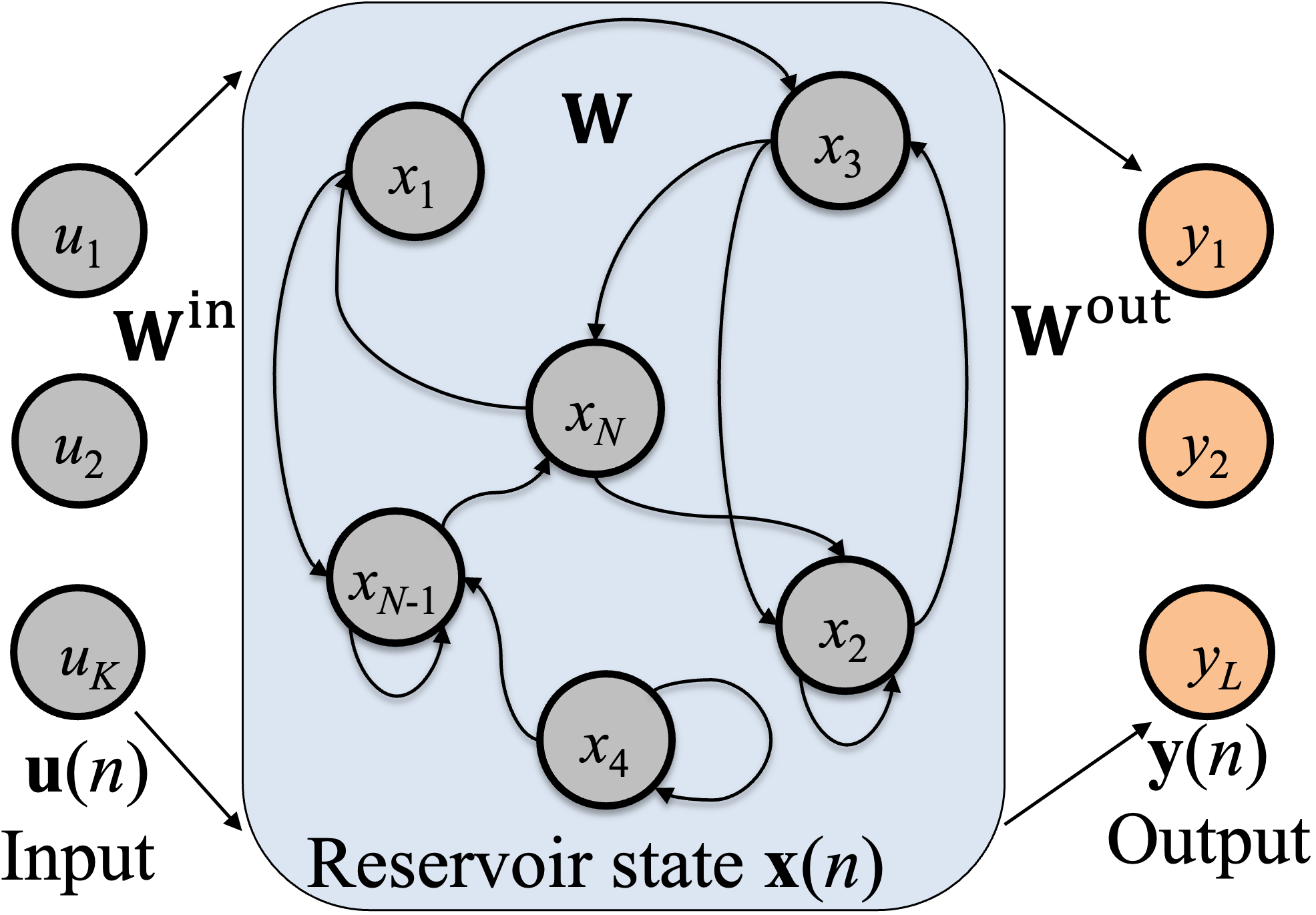}
\caption{Standard architecture of an echo state network. 
The current input to the network $\mathbf{u}(n)$  is projected to the reservoir with $N$ neurons using a random matrix $\mathbf{W}^{\mathrm{in}}$.
The neurons within the reservoir are interconnected through another random matrix $\mathbf{W}$.
The trainable readout network with weight matrix $\mathbf{W}^{\mathrm{out}}$ computes the output $\mathbf{y}(n)$ from the current reservoir state $\mathbf{x}(n)$.
}
\label{fig:esn}
\end{figure}
    
\section{Introduction}
\label{sect:intro}

Reservoir computing~\cite{RC09, yan2024emerging} is a framework for processing spatio-temporal input patterns in a neural network that consists of three stages depicted in Fig.~\ref{fig:esn}. 
The central stage is a recurrent neural network with fixed dynamics -- {\it the reservoir} -- that receives input through a fixed feed-forward matrix, the {\it input projection matrix}. The reservoir feeds into the third stage, the {\it readout network}, consisting of the {\it readout matrix} and a layer of output neurons, a feedforward network which often is trained in a supervised fashion.
The sustained dynamic state of the reservoir acts like a working memory, storing past and present inputs. The readout matrix can decode this information, to perform tasks such as memorization, time-series prediction or classification. 

The reservoir can consist of a dynamical system with fixed parameters of any type that retains information about past inputs. For example, the reservoir 
can be a network of spiking neurons like in liquid state machines~\cite{MaassLSM2002}, consist of neurons with quantized states as in neural prediction machines~\cite{tino2004markovian}, or even have recurrent weights with short-term plasticity, like in state-dependent networks~\cite{BuonomanoTemporal1995}.
Here we focus on the most common model for reservoir computing, echo state networks (ESN)~\cite{JaegerEcho2001}, in which the reservoir is an artificial neural network with a fixed recurrent random matrix, and the projection matrix is also random. 
ESN models differ not only in the type of recurrent neural network used as reservoir, but also in how the weights of the readout network are formed. In many ESN models, the readout matrix is optimized by supervised learning (i.e., teacher forcing~\cite{ESNtut12}), which is feasible if sufficient training data are available. 
In other types of ESNs, the readout matrix can be directly computed from the weights of the reservoir and input projection matrices, without any training~\cite{FradyNeCo2018}.

The critical performance measure for a reservoir computing network is its memory capacity~\cite{JaegerMemory2002}, which describes the maximal amount of information that can be contained within the current state. 
However, memory capacity has only been estimated for simple ESN models, such as those with linear neurons and orthogonal recurrent matrices~\cite{WhiteOrthogonal2004, GanguliMemory2008}.
The goal of this article is to develop a theory to estimate the memory capacity for a large variety of ESN models, (Section~\ref{sec:capacity:applicability}), in particular, including those with nonlinear reservoir neurons.

Two earlier studies of ESNs~\cite{FradyNeCo2018,KleykoPerceptron2020} used the framework of hyperdimensional computing or vector symbolic architectures~\cite{KleykoSurveyVSA2021Part1,KleykoSurveyVSA2021Part2} to dissect the computations in the reservoir into two parts: adding time tags to individual inputs, and superimposing them into the reservoir state. Using this dissection, it is possible to precompute the readout matrices, thereby avoiding the need of training. 
Further, these studies proposed using the statistics of inner products between the reservoir state and readout matrices for estimating the accuracy of the ESN. This type of ESN theory has been called the \textit{perceptron theory} because the ESN readout network is identical to the earliest and extensively studied neural network, the
Perceptron~\cite{rosenblatt1958perceptron,minkypappert,amari1977competition,grossberg1988nonlinear,maass2000computational,steinbuch1963learning,PetersonEtAl1954}.
Here we demonstrate experimentally that the most general form of the perceptron theory can accurately predict the performance of a large variety of ESNs.

The article is structured as follows. 
The methods and the perceptron theory for understanding the capacity of ESNs are presented in Section~\ref{sect:methods}. 
The evaluation of the theory, based on comparisons to empirical ESN accuracy, follows in Section~\ref{sect:restuls}. 
Finally, Section~\ref{sect:discussion} presents a discussion, related work, and concluding remarks.


\section{Methods}
\label{sect:methods}

\subsection{Trajectory association task and its modifications}
\label{sect:methods:trajectory}

We consider the trajectory association task~\cite{PlateRecurrent1992}, which is commonly used to assess working memory capacity in recurrent neural networks.
The trajectory association task involves sequences of symbols $\mathbf{s}$ that are generated at random. The symbols are selected from an alphabet of size $D$ and can be represented as $D$-dimensional one-hot vectors. Distributed bipolar representations of the symbol representations with small correlations can be modeled by multiplying the orthogonal one-hot encodings of symbols with a codebook matrix $\mathbf{\Phi}  \in \{-1,1\}^{N \times D}$, a matrix with dense bipolar random entries.  At each position $n$ in a sequence, a symbol vector is chosen with uniform probability. Trajectory association includes two phases, memorization and recall. During {\it memorization}, the sequence is sent, 
a symbol $\mathbf{s}(n)$ at each timestep, to a working memory. During {\it recall}, while the memory continues to receive new inputs, the memory state at timestep $n$ is used to decode the symbol presented $d$ steps ago (i.e., at timestep $n-d$), where the integer $d$ denotes the delay. Naturally, networks with higher memory capacity will be able to correctly recall symbols up to larger values of delay.
The total length of the sequence $\mathbf{s}$ is $E+M+R$; the first $E+M$ timesteps correspond to the memorization phase. The first $E$ reservoir states of the network (usually, we set $E$ to $1,000$ in the experiments below) are not tested for recall, they serve to get the network from the initial empty state to a steady state behavior (often called warm-up).
The subsequent $M$ inputs are used to train the network (teacher forcing phase) while the last $R$ states (recall phase) are used to assess network's performance. 

The mapping of trajectory association to reservoir computing is straightforward. The inputs, $\mathbf{u}(n)$, are the orthogonal one-hot representations of symbols in a sequence, the projection matrix is the codebook, $\mathbf{W}^{\mathrm{in}} = \mathbf{\Phi}$, the reservoir is the working memory, and the output stage is used to decode past inputs from a reservoir state.

\paragraph*{\textit{n}-Bit Memory Task} Another symbolic memorization task, popular for recurrent neural networks, is the $n$-Bit Memory Task~\cite{JaegerMemoryESN2012}.
In this task, there is a signal composed of a sequence of symbols, which is followed by a series of $T$ distractor symbols and a final cue symbol. Once the network receives the cue symbol, it should output the entire sequence of symbols.
The task is completed successfully if all the signal symbols are retrieved correctly. 
In previous work~\cite{JaegerMemoryESN2012}, typical values of $n$ are 5 (2 symbols, sequence length 5) or $\approx23$ (5 symbols, sequence length 10).

If the cue symbol is presented to the network before the distractor sequence, and the delay, $d$ is equal to $T$, then the $n$-Bit Memory Task can be cast as a trajectory association task. That is because this modification will result in \textit{retrieval} of the signal during the distractor sequence, and the cue symbol after the last distractor symbol. Therefore, our main results focus on trajectory association, and we report experiments using the modified formulation of the $n$-Bit Memory task in Supplementary Note~\ref{sect:perf:5:bit}. Finally, we note that this problem formulation also highly resembles the episodic copy task that is commonly used in evaluating recurrent neural networks~\cite{ArjovskyEvolutionRNN2016,DanihelkaAssociative2016}.

\subsection{Echo state networks}
\label{sect:esn}

Following~\cite{ESNtut12}, we formally define the core components of ESNs (Fig.~\ref{fig:esn}), as well as the underlying reservoir dynamics. 
The projection matrix $\mathbf{W}^{\mathrm{in}}$ is a fixed matrix that projects the $K$-dimensional input patterns $\mathbf{u}(n)$ to the reservoir.
The reservoir consists of $N$ neurons, recurrently interconnected by a fixed matrix 
$\mathbf{W}$. The matrices $\mathbf{W}^{\mathrm{in}}$ and $\mathbf{W}$ are usually generated at random\footnote{There are, however, variants where $\mathbf{W}$ is initialized in a structured way, see, e.g.,~\cite{RodanMinESN2011,intESN}. 
Below, we will consider one such variant.} during the network initialization. 
The readout network contains a matrix $\mathbf{W}^{\mathrm{out}}$ that transforms a state of the reservoir at a timestep, $\mathbf{x}(n)$, into an $L$-dimensional output state $\mathbf{y}(n)$.     

\subsubsection{Reservoir}
In its most general form, the reservoir is a recurrent neural network whose state $\mathbf{x}(n)$ evolves according to the following discrete time dynamics:
\noindent
\begin{equation}
\mathbf{x}(n) \!=\! (1\!-\!\alpha)\mathbf{x}(n\!-\!1) \!+\! \alpha \tanh(\gamma\mathbf{W}\mathbf{x}(n\!-\!1) \!+\! \beta\mathbf{W}^{\mathrm{in}}\mathbf{u}(n)).
\label{eq:esnres}
 \end{equation}
 \noindent
Here, $\mathbf{u}(n)$ is the input at timestep $n$, and $\tanh()$ is the transfer function of the reservoir neurons.
The choice of the squashing nonlinearity $tanh()$ as the transfer function limits the activities of reservoir neurons to the range $[-1, 1]$, preventing the activity from exploding.
In addition, there are three hyperparameters:
$\alpha \in [0,1]$ controls the relative influence of the previous state versus the new inputs and feedback.
$\beta$ and $\gamma$ scale the input and the recurrent feedback, respectively.
Particular settings of the hyperparameters correspond to special variants in which some terms of Eq.~(\ref{eq:esnres}) are removed, for example, for ESN without leaky integration: $\alpha =1$. In addition, we will also consider the case (Eqs.~(\ref{eq:esnres:var1}) \&~(\ref{eq:esnres:var2})) of reservoir neurons with a linear transfer function.

A key desideratum for ESNs is the eponymous echo state property: ``the state of the reservoir $\mathbf{x}(n)$ should be uniquely defined by the fading history of the input $\mathbf{u}(n)$''~\cite{ESNtut12}. For ESNs, this property is achieved by ensuring the spectral radius of the recurrent matrix $\mathbf{W}$ is less than $1$. The spectral radius of a matrix is the maximum of the absolute values of its eigenvalues. Therefore, some form of normalization to random matrices is common in practice.
Here, we study two types of the recurrent matrix: a) ring-based reservoir~\cite{RodanMinESN2011,intESN} when $\mathbf{W}$ is a permutation matrix and b) when $\mathbf{W}$ is a random orthogonal matrix formed by applying the QR decomposition to a random matrix generated from the standard normal distribution.
For both types, the spectral radius of $\mathbf{W}$ is exactly one and the feedback can be made contracting by choosing $\gamma<1$.

\subsubsection{Readout network}
\label{sect:readout}

The readout network computes the output of the ESN from the reservoir state as:
\noindent
\begin{equation}
\mathbf{y}(n)=g\left(\mathbf{W}^{\mathrm{out}}\mathbf{x}(n)\right).
\label{eq:esny}
\end{equation}
\noindent
Here, $\mathbf{W}^{\mathrm{out}}$ is the readout matrix and $g()$ is the transfer function of the output neurons. 
In the following experiments, we use the winner-take-all function,
 $\argmax(\cdot)$, followed by the indicator function, $\mathbbm{1}(\cdot)$: $g(\mathbf{x})= \mathbbm{1} (\argmax ( \mathbf{x}))$.

For the trajectory association task, the output vector $\mathbf{y}(n)$ is a $D$-dimensional one-hot vector representing a symbol $\mathbf{s}(n-d)$ from the input sequence.
The readout matrix for delay $d$, $\mathbf{W}^{\mathrm{out}}(d)$, the ESN estimate of the past input symbol is: 
\noindent
\begin{equation}
\hat{\mathbf{s}}(n-d)= \mathbbm{1}_D \left( \argmax \left( \mathbf{W}^{\mathrm{out}}(d) \mathbf{x}(n)  \right)  \right),
\label{eq:recall}
\end{equation}
\noindent
where $\mathbbm{1}_D(i)$ returns a $D$-dimensional one-hot vector with $1$ in component $i$.
The only parameters in the ESN that can be tuned to minimize the error of the estimate Eq.~(\ref{eq:recall}) are the 
readout matrices $\mathbf{W}^{\mathrm{out}}(d)$. 
Two commonly used methods for adjusting the readout matrix are described below, and a new method based on the covariance matrix is presented in Section~\ref{sect:restuls:covariance}.

\paragraph{Readout through linear regression}
\label{sect:readout:regression}

A common approach in ESNs is to form the readout matrix with supervised learning based on linear regression~\cite{ESNtut12}. 
In essence, the readout matrix is optimized as a regressor between the reservoir states and one-hot encodings of a teacher signal~\cite{williams1989learning}, providing the correct past input symbols: 
\begin{equation}
\mathbf{W}^{\mathrm{out}}(d) = \mathbf{Y}^{\top}(d) \mathbf{X}
\left ( \mathbf{X}^{\top} \mathbf{X} + \lambda \mathbf{I} \right )^{-1}.
\label{eq:readout:regr}
\end{equation}
\noindent 
Here, the matrix $\mathbf{X} \in [M \times N]$ contains $M$ reservoir states $\mathbf{x}(n)$ during the memorization phase ($n \in [E+1, E+M]$), the matrix $\mathbf{Y}(d) \in [M \times D]$ stores the corresponding one-hot encodings of symbols $\mathbf{s}(n)$ at a given delay value $d$, i.e., $n \in [E+1-d, E+M-d]$, and $\mathbf{I} \in [N \times N]$ is the identity matrix;  $\lambda$ is the regularization parameter of ridge regression.

The advantage of the readout through linear regression is that the method is generic and can be applied across many different types of tasks, such as classification, detection, identification, and prediction.

\paragraph{Readout without learning}
\label{sect:readout:codebook}
The simplest form of readout in ESNs uses ideas from
the hyperdimensional computing framework~\cite{Gayler2003, Kanerva09, FradySDR2020, KleykoFramework2021}.
The key idea is that the codebook in the fixed input matrix $\mathbf{W}^{\mathrm{in}} = \mathbf{\Phi}$ can be used directly to measure the similarity (inner product) between representations of the symbols stored in the reservoir. For the most recent symbol added to the reservoir ($d=0$), the readout matrix is: $\mathbf{W}^{\mathrm{out}}(0)= \mathbf{\Phi}^\top$. 
For positive delays $d>0$, one has to account for the transformation by repeated passes through the recurrent matrix $\mathbf{W}$. Thus, in general, the readout matrix can be written as:
\begin{equation}
\mathbf{W}^{\mathrm{out}}(d)= \mathbf{\Phi}^\top 
\mathbf{W}
^{-d}.
\label{eq:readout:VSA}
\end{equation}
\noindent
The advantage of this approach is that no training is required to obtain the readout matrix.

\subsection{Analyzing ESN networks}
\label{sec:capacity}

It is easy to realize~\cite{FradyNeCo2018} that the readout network of the ESN, Eq.~(\ref{eq:recall}), is a one-layer perceptron~\cite{rosenblatt1958perceptron}, in which the input, the reservoir state, is multiplied by the readout matrix and the largest vector component extracted by a nonlinear winner-take-all (WTA) mechanism.  In its long history, the WTA perceptron~\cite{steinbuch1963learning} has appeared under various names, such as  linear machine~\cite{nilsson1965learning}, or winner-take-all group~\cite{gallant1993winner}, and has been subject to multiple approaches of analysis.

\subsubsection{Theory of the WTA perceptron}
\label{sec:capacity:perceptron}

The function of WTA perceptrons can be analyzed by a generalized form of signal detection theory~\cite{mcnicol2005primer}, an approach partially described in earlier papers \cite{joseph1961contributions, babadi2014sparseness, FradyNeCo2018, KleykoPerceptron2020}. Each output neuron represents a different symbol, and a theory should predict the accuracies with which each symbol is produced in the output. In the context of ESNs, the essential statistics for predicting accuracy are the inner products between reservoir state and rows of the readout matrix. 
The response is correct if the inner product of the neuron corresponding to the symbol of the input encoded by the reservoir is larger than for all other $D-1$ ``distractor'' symbols. The most general form of WTA perceptron theory~\cite{KleykoPerceptron2020} estimates the probability of the correct response, i.e., the accuracy, for input symbol $i=1$ as: 
\begin{equation}
\begin{split}
p_1:&=
p( \hat{\mathbf{s}} = \mathbbm{1}_D(1) |  \mathbf{s} = \mathbbm{1}_D(1)  )= \\
&=\int_{-\infty}^{\infty} dh_1
\int_{-\infty}^{h_1} dh_2
\dots 
\int_{-\infty}^{h_1} dh_D
\; p(\mathbf{h}),
\end{split}
 \label{eq:pcorr:mvn}
 \end{equation}
\noindent
Here, the multivariate distribution $p(\mathbf{h})$ describes the {\it input statistics of the output neurons}, i.e., the inner products between reservoir state and rows of the readout matrix, $h_j := \mathbf{W}^{\mathrm{out}}_{j:}(d)^{\top} \mathbf{x}(n); j=1,..., D$.

To aggregate the accuracy of all different symbols into the overall expected accuracy, one forms the expectation over all symbols:
$\sum_{i=1}^{D} f_i p_i$, where $f_i$ is the prior probability of the $i$th symbol in the data. To assess the correctness of the predictions provided by the theory, the expected accuracy can be compared to the empirically observed one.

\subsubsection{Approximations of the full theory}
\label{sec:capacity:perceptron:approx}

The theory of the WTA perceptron in Eq.~(\ref{eq:pcorr:mvn}) can only be evaluated if there is a concrete expression for the multivariate distribution $p(\textbf{h})$, which typically relies on simplifying assumptions~\cite{KleykoPerceptron2020}. The weakest commonly made assumption is that the distribution is a multivariate Gaussian, $p(\mathbf{h})={\cal N}(\mathbf{h}, \bm{\mu}, \bm{\Sigma})$,  
where $\bm{\mu}$ and $\bm{\Sigma}$ denote the first two moments, the mean vector and the covariance matrix, respectively. Eq.~(\ref{eq:pcorr:mvn}) with the assumption of normality is the most general tractable approximation of the WTA perceptron theory that we will explore here.

If one further assumes that the Gaussian distribution factorizes, i.e., the activations of output neurons 
are statistically independent, 
Eq.~(\ref{eq:pcorr:mvn}) simplifies to a single integral~\cite{FradyNeCo2018}:
\noindent
\begin{equation}
p_1=\int_{-\infty}^{\infty} \frac{dh}{\sqrt{2 \pi} \bm{\sigma}_{i}} e^{-\frac{ (h-\bm{\mu}_{i})^2 }{2 \bm{\sigma}_{i}^2}} \prod_{j=1, j \neq i}^{D} \Phi(h,\bm{\mu}_{j}, \bm{\sigma}_{j}). 
 \label{eq:pcorr:indep}
 \end{equation}
\noindent
where $\Phi(\cdot)$ is the cumulative Gaussian and $\bm{\sigma}$ denotes the vector of standard deviations of individual inner products.


If, in addition, one assumes that the Gaussian distributions of all ``distractor'' symbols are not only independent but also identical, Eq.~(\ref{eq:pcorr:indep}) becomes~\cite{FradyNeCo2018}: 
\noindent
\begin{equation}
p_1 =\int_{-\infty}^{\infty} \frac{dh}{\sqrt{2\pi}}  \; e^{-\frac{1}{2}{h}^2} \left[ \Phi\left(\frac{\sigma_{r}}{\sigma_{h}}h+\frac{\mu_{h}-\mu_{r}}{\sigma_{h}}\right)\right]^{D-1}.
 \label{eq:pcorr:orig1}
 \end{equation}
 \noindent
Here,  $\mu_{h}$ and $\sigma_{h}$ denote mean and standard deviation of the input to the output neuron that corresponds to the correct symbol (hit);  $\mu_{r}$ and $\sigma_{r}$ denote the mean and standard deviation of the inner products for all other distractor neurons (reject). 

Fig.~\ref{fig:ESN:acc:perc} in Supplementary Note~\ref{sec:theory:comparisons} reports the comparison of different approximation levels of the theory, Eqs.~(\ref{eq:pcorr:mvn})-(\ref{eq:pcorr:orig1}).


\begin{table}[tb]
\renewcommand{\arraystretch}{1.0}
\caption{Special cases of reservoir dynamics, Eq.~(\ref{eq:esnres})\\
}
\label{tab:ESN:var}
\begin{minipage}[h]{1.0\linewidth}
    \begin{center}
    \begin{tabular}{c|c|c|c|c}\hline
    \multicolumn{4}{c|}{\textbf{Reservoir specification}} & \multirow{2}{*}{\makecell{\textbf{Reservoir} \\  \textbf{dynamics}}} \\  \cline{1-4} 
     \makecell{Transfer \\ function}  &  \makecell{Input \\ scaling, $\beta$}  & \makecell{Recurrent \\ decay, $\gamma$}  & Leakage, $\alpha$  \\ \hline 
     \multirow{2}{*}{\rotatebox[origin=c]{90}{lin.}} & \xmark & \xmark  & \xmark & Eq.~(\ref{eq:esnres:var1}) \\ \cline{2-5} 
     & \cmark & \cmark & \xmark & Eq.~(\ref{eq:esnres:var2})  \\ \hline    
    \multirow{3}{*}{\rotatebox[origin=c]{90}{$\tanh$}} & \cmark & \xmark  & \xmark & Eq.~(\ref{eq:esnres:var3})  \\ \cline{2-5} 
     & \cmark & \cmark  & \xmark  & Eq.~(\ref{eq:esnres:var4}) \\ \cline{2-5}          
     & \cmark & \cmark  & \cmark & Eq.~(\ref{eq:esnres:var5}) \\ \hline   
    \end{tabular}
    \end{center}
\end{minipage}
\vfill
\begin{minipage}[h]{1.0\linewidth}
\begin{eqnarray}
\mathbf{x}(n)&=&\mathbf{W}\mathbf{x}(n-1) + \mathbf{W}^{\mathrm{in}}\mathbf{u}(n)
\label{eq:esnres:var1}\\
\mathbf{x}(n)&=&\gamma\mathbf{W}\mathbf{x}(n-1) + \beta\mathbf{W}^{\mathrm{in}}\mathbf{u}(n)
\label{eq:esnres:var2}\\
\mathbf{x}(n)&=&\tanh(\mathbf{W}\mathbf{x}(n-1) + \beta\mathbf{W}^{\mathrm{in}}\mathbf{u}(n))
\label{eq:esnres:var3}\\
\mathbf{x}(n)&=&\tanh(\gamma\mathbf{W}\mathbf{x}(n-1) + \beta\mathbf{W}^{\mathrm{in}}\mathbf{u}(n))
\label{eq:esnres:var4}\\
\mathbf{x}(n)&\!\!=\!\!&(1\!\!-\!\!\alpha)\mathbf{x}(n\!\!-\!\!1)\!\!+\!\!\alpha \! \tanh(\! \gamma \! \mathbf{W}\mathbf{x}(n\!\!-\!\!1)\!\!+\!\!\beta \!\mathbf{W}^{\mathrm{in}}\mathbf{u}(n))
\label{eq:esnres:var5}
\end{eqnarray}
\end{minipage}
\end{table}

\section{Results}
\label{sect:restuls}

\subsection{Covariance-based readout without learning}
\label{sect:restuls:covariance}

Optimizing readout matrices in ESNs without requiring supervised learning (Section~\ref{sect:readout:codebook}) straightforward advantages in efficiency. 
However, Eq.~(\ref{eq:readout:VSA}) yields a suboptimal readout matrix, unless the columns of $[\mathbf{\Phi},\mathbf{W}^1 \mathbf{\Phi}, \mathbf{W}^2 \mathbf{\Phi}, \ldots, \mathbf{W}^n \mathbf{\Phi}]$ are perfectly orthogonal.
Here we present two novel ways to optimize the readout for trajectory association that work under more general conditions by including the appropriate covariance estimates.  
These novel ESN models without learning also provide additional insights into ESNs with regression-based learning (see Section~\ref{sec:results:readouts:structure}). 

The codebook-based readout Eq.~(\ref{eq:readout:VSA}) can be generalized to non-orthogonal reservoir states by multiplying with the inverse of an estimate of the covariance matrix of the reservoir states (denoted as $\tilde{\mathbf{C}}$): 
\noindent
\begin{equation}
\mathbf{W}^{\mathrm{out}}(d)= \mathbf{\Phi}^\top 
\mathbf{W} 
^{d} \tilde{\mathbf{C}}^{-1}.
\label{eq:readout:cov}
\end{equation}
\noindent
We use two estimates of $\tilde{\mathbf{C}}$: a coarse and a refined estimate. 
The coarse estimate is: 
\noindent
\begin{equation}
\tilde{\mathbf{C}}= \sum_{n=1}^{\infty} \gamma ^{2n} \mathbf{W}^n \mathbf{\Phi} \left( \frac{1}{D}\mathbf{I} \right) \mathbf{\Phi}^\top \mathbf{W}^{-n},
\label{eq:cov:coarse:gamma}
\end{equation}
\noindent
where $\mathbf{I} \in [D \times D]$ is the identity matrix.
The refined estimate is: 
\noindent
\begin{equation}
\begin{split}
&\tilde{\mathbf{C}} = \sum_{n=1}^{\infty} \gamma ^{2n} \mathbf{W}^n \mathbf{\Phi} \left( \frac{1}{D}\mathbf{I} \right) \mathbf{\Phi}^\top \mathbf{W}^{-n} + \\
&+\sum_{n_1=1}^{\infty} \sum_{n_2=1,n_2\neq n_1}^{\infty} \gamma ^{n_1 + n_2} \mathbf{W}^{n_1} \mathbf{\Phi} \left( \frac{1}{D^2}\mathbf{J} \right) \mathbf{\Phi}^\top \mathbf{W}^{-n_2},
\end{split}
\label{eq:cov:fine:gamma}
\end{equation}
\noindent
where $\mathbf{J} \in [D \times D]$ is the all-ones matrix.
Note, however that these estimates of $\tilde{\mathbf{C}}$ apply only for reservoir neurons with a linear transfer function and the hyperparameter setting $\alpha=1$, (cf. Eqs.~(\ref{eq:esnres:var1}) \&.~(\ref{eq:esnres:var2})).

Like in the codebook-based readout, the matrices of this novel covariance-based readout can be pre-computed from $\mathbf{W}$ and $\mathbf{\Phi}$, and no learning is required. Evidently, the construction of the readout matrix is computationally more expensive with the refined versus with the coarse covariance estimate.

\subsection{Overview over applicability of the perceptron theory}
\label{sec:capacity:applicability}

\begin{table}[tb]
\renewcommand{\arraystretch}{1.0}
\caption{Applicability of the perceptron theory to different ESN models
}
\label{tab:ESN:theory}
    \begin{center}
    \begin{tabular}{c|c|c|c?c|c|c?c|c|c}\cline{2-10}  
    \multicolumn{1}{c}{} & \multicolumn{3}{c?}{Codebook, (\ref{eq:readout:VSA})} & \multicolumn{3}{c?}{Regression, (\ref{eq:readout:regr})} & \multicolumn{3}{c}{Cov. estimator, (\ref{eq:readout:cov})} \\ \cline{2-10} 
    \multicolumn{1}{c}{} &  \textbf{Q1} & \textbf{Q2} & \textbf{Q3} & \textbf{Q1} & \textbf{Q2} & \textbf{Q3} & \textbf{Q1} & \textbf{Q2} & \textbf{Q3} \\ \cline{2-10} \hline 
    \multirow{5}{*}{\rotatebox[origin=c]{90}{Permutation}}  
     &  \cmark & \cmark & \cmark & \xmark & \cmark  & \xmark & \cmark & \cmark  & \xmark \\ \cline{2-10}  
     &  \cmark & \cmark & \cmark & \xmark & \cmark  & \xmark & \cmark & \cmark  & \xmark \\ \cline{2-10}   
     &  \cmark & \cmark & \cmark & \xmark & \cmark  & \xmark & \xmark  & \xmark  & \xmark  \\ \cline{2-10}
     &  \cmark & \cmark & \cmark & \xmark & \cmark  & \xmark & \xmark  & \xmark  & \xmark  \\ \cline{2-10}         
     &  \cmark & \cmark & \cmark & \xmark & \cmark  & \xmark & \xmark  & \xmark  & \xmark  \\ \hline  \hline 

    \multirow{5}{*}{\rotatebox[origin=c]{90}{ Rand. ortn.}} 
     &  \cmark & \cmark & \cmark & \xmark & \cmark  & \xmark & \cmark & \cmark  & \xmark \\ \cline{2-10}    
     &  \cmark & \cmark & \cmark & \xmark & \cmark  & \xmark & \cmark & \cmark  & \xmark \\ \cline{2-10}   
     &  \cmark & \cmark  & \xmark & \xmark & \cmark  & \xmark & \xmark  & \xmark  & \xmark  \\ \cline{2-10}     
     &  \cmark & \cmark  & \xmark & \xmark & \cmark  & \xmark & \xmark  & \xmark  & \xmark  \\ \cline{2-10}     
     &  \cmark & \cmark  & \xmark & \xmark & \cmark  & \xmark & \xmark  & \xmark  & \xmark  \\ \hline    
    
    \end{tabular}
    \end{center}
\end{table}

In essence, perceptron theory estimates the statistics of the inner products between reservoir states and rows in the readout matrix. The statistics of reservoir states critically depends on the projection matrix, choices of hyperparameters, and the transfer function in Eq.~(\ref{eq:esnres}). 
Table~\ref{tab:ESN:var} summarizes the five model configurations of ESNs that are explored in our experiments.
Each row in the table specifies a single model variant. A cross mark \xmark $\,$  indicates that the corresponding parameter is set to one 
and not adjustable. 
A tick mark \cmark $\,$ indicates that the parameter is tunable. Roughly, the earlier rows correspond to simpler models, and the models in later rows have higher complexity.

The rows of the table correspond to the equations Eqs.~(\ref{eq:esnres:var1})-(\ref{eq:esnres:var5}) governing the reservoir dynamics.
Note that the last row corresponds to the reservoir dynamics in Eq.~(\ref{eq:esnres}).
For example, in the simplest model described by Eq.~(\ref{eq:esnres:var1}), the reservoir neurons are linear and $\gamma=1$, thus, there is no recency effect, and all inputs $\mathbf{u}(n)$ are given equal priority.

Table~\ref{tab:ESN:theory} summarizes which of the ESN variants can be optimized and analyzed with the perceptron theory presented here. 
It answers the following three questions, with green tick marks and red cross marks indicating ``yes'' and ``no'', respectively:
\noindent
\begin{itemize}
     \item[\textbf{Q1}:] Given the reservoir dynamics, can the readout network be optimized without learning?
    \item[\textbf{Q2}:] Given the input statistics in the output neurons, can our theory predict the accuracy?    
     \item[\textbf{Q3}:] Given just the model hyperparameters, can our theory predict the accuracy?
\end{itemize}
\noindent
The answers in Table~\ref{tab:ESN:theory} summarize results from theoretical analysis and from experiments. 
Specifically, the answers to \textbf{Q1} rely on the results in Sections~\ref{sect:readout} and~\ref{sect:restuls:covariance}. The answers to \textbf{Q3} are based on the theoretical results from~\cite{FradyNeCo2018}. The answers to \textbf{Q2} combine the \textbf{Q3} results with experimental observations detailed in the following Section~\ref{sect:perf}.

Table~\ref{tab:ESN:theory} covers $30$ ($5 \times 2 \times 3$) different ESN models. 
The upper half contains models where the reservoir connectivity matrix $\mathbf{W}$ is a random permutation; the lower half contains models where $\mathbf{W}$ is a random orthogonal matrix. Individual rows in the upper and lower parts of Table~\ref{tab:ESN:theory} correspond to the reservoir dynamic equations specified by the rows of Table~\ref{tab:ESN:var}.
The three sections of table columns cover the three methods to construct the readout matrix in an ESN described in Sections~\ref{sect:readout} and~\ref{sect:restuls:covariance}. 

In summary, the $\textbf{Q1}$ answers show that readout matrices can be optimized without learning for all codebook-based estimators, as well as for the covariance-based estimators, when the reservoir is linear. 
Answers to $\textbf{Q2}$ suggest that the perceptron theory can predict the performance of any ESN model, if the statistics of inner products (mean values and covariance matrices) is available.\footnote{
Note that the answers to $\textbf{Q2}$ for the case of the covariance-based readout matrix are negative only when it is not possible to calculate the readout.
} 
In contrast, answers to $\textbf{Q3}$ indicate that
predicting the ESN performance solely from the hyperparameters of the ESN is challenging.
Nevertheless, the techniques developed so far allow doing so for some of the ESN variants when the readout matrix is codebook-based.
Specifically, our theory only succeeds for all five models when $\mathbf{W}$ is a random permutation matrix; it also succeeds for two models with the linear transfer function when $\mathbf{W}$ is a random orthogonal matrix.

\subsection{Predicted and experimental recall accuracies}
\label{sect:perf}

Here we describe experiments and theoretical predictions of how well information from  different variants of ESN can be recalled.
At the recall phase, the ESN uses the state of its reservoir $\mathbf{x}(n)$ as the query vector to retrieve the symbol stored $d$ steps ago, where $d$ denotes delay. 
In the experiments, the range of the delays varied between $0$ and $25$.
The recall is performed with the readout matrix formed for a particular value of $d$; this matrix has one $N$-dimensional vector per each symbol.

\subsubsection{Predicting accuracies of ESNs with codebook-based readout matrix}
\label{sect:perf:analytical}

For the codebook-based readout Eq.~(\ref{eq:readout:VSA}), and see Table~\ref{tab:ESN:theory}, one can compute the statistical moments required by the perception theory, Eq.~(\ref{eq:pcorr:orig1}), directly from the model hyperparameters.

\paragraph{Linear reservoir dynamics}

\begin{figure}[tb]
\centering
\includegraphics[width=1.0\linewidth]{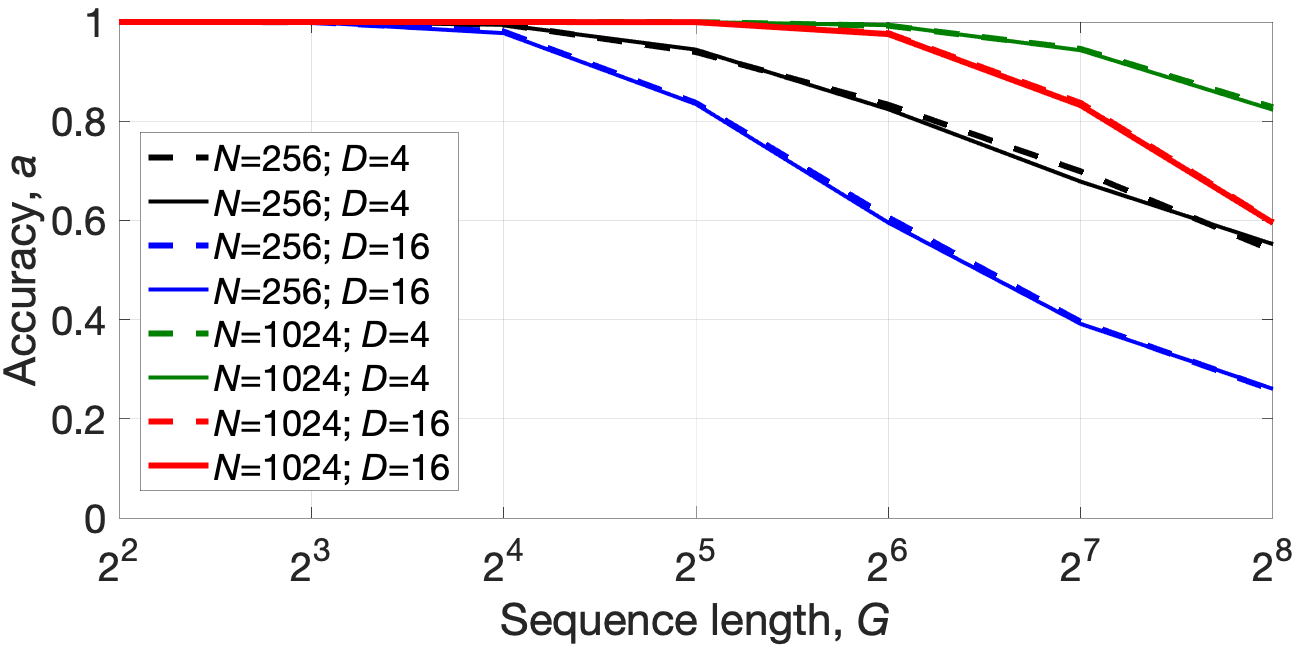}
\caption{
Predicted (solid lines) and experimental (dashed lines) recall accuracies for different lengths of memorized sequences when the reservoir was updated according to Eq.~(\ref{eq:esnres:var1}); $\mathbf{W}$ was a random permutation matrix but the results are the same when $\mathbf{W}$ is a random orthogonal matrix; $\mathbf{W}^{\mathrm{out}}(d)$ was the codebook-based readout matrix.
Eq.~(\ref{eq:pcorr:orig1}) was used to obtain analytical results. 
$N$ was in $\{256, 1024\}$; $D$ was in $\{4, 16\}$.
The empirical results were averaged over $10$ simulation runs. 
Each simulation used $128$ random sequences to estimate the recall accuracy. 
}
\label{fig:ESN:accuracy:linear}
\end{figure}

\begin{figure}[b]
\centering
\includegraphics[width=1.0\linewidth]{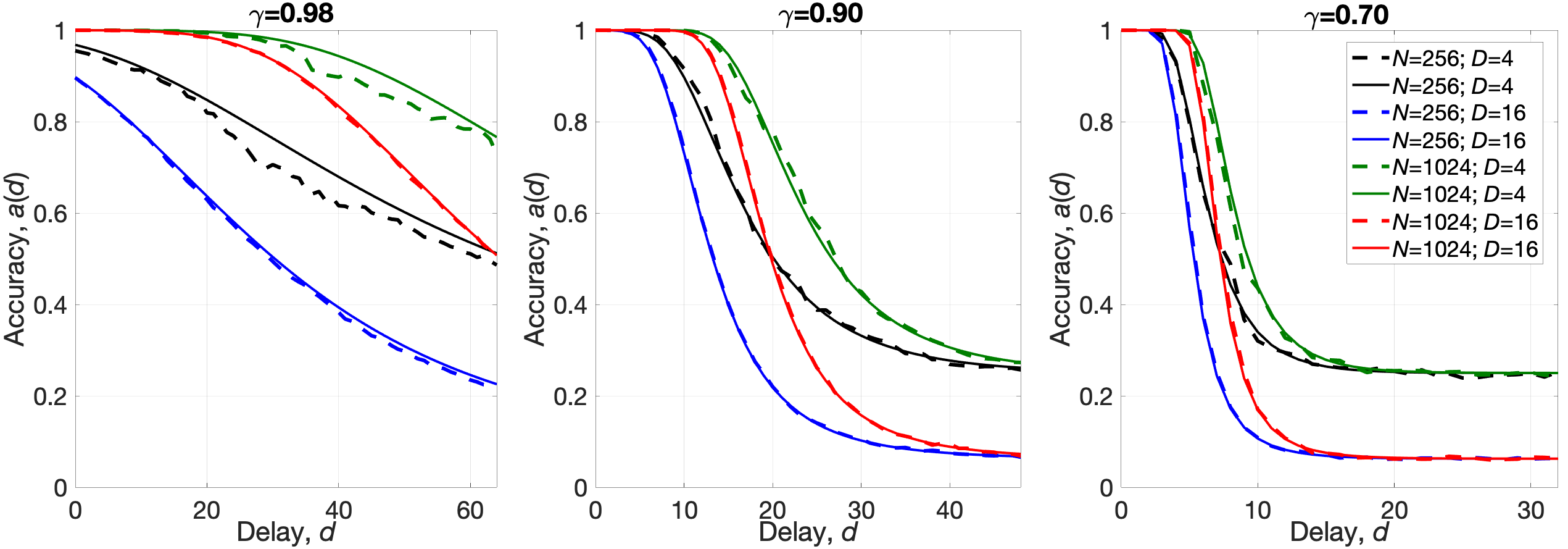}
\caption{
Predicted (solid lines) and experimental (dashed lines) recall accuracies for the linear reservoir with decay $\gamma$, Eq.~(\ref{eq:esnres:var2}); $\mathbf{W}$ was a random permutation matrix; $\mathbf{W}^{\mathrm{out}}(d)$ was the codebook-based readout matrix.
Eq.~(\ref{eq:pcorr:orig1}) was used to obtain predicted accuracies.
$N$ was in $\{256, 1024\}$; $D$ was in $\{4, 16\}$.
The empirical results were averaged over $10$ simulation runs. Each simulation used a random sequence with $E=1,000$, $M=0$, and $R=3,000$. 
}
\label{fig:ESN:accuracy:gamma}
\end{figure}

If the transfer function of reservoir neurons is linear (and the recurrent matrix has unit eigenvalues), as in reservoir update Eq.~(\ref{eq:esnres:var1}), there is no recency effect. Thus, the recall accuracy is the same for all $d$ as the traces of previous symbols are not decaying.  
Of course, the recall accuracy will still decay with the length of the sequence $\mathbf{s}$.
To explore this length effect efficiently, we slightly modify the  setup for the trajectory association task.
Specifically, we generate many random sequences with some fixed length $G$, not just one. A subset of these sequences will be used to train the readout matrix while the compliment will be used to estimate the recall accuracy. Note, that in our notation of the trajectory association task, the sequences used for training have $E=0$, $M=G$, and $R=0$, while for those used for recall have $E=0$, $M=0$, and $R=G$. 
The accuracy is then tested for the final state of the network $\mathbf{x}(G)$ for both subsets.

Fig.~\ref{fig:ESN:accuracy:linear} presents analytical and empirical recall accuracies as a function of the length of the memorized sequence $G$ for four combinations of $N \in \{256, 1024\}$ and $D \in \{4, 16\}$ for the reservoir updated with Eq.~(\ref{eq:esnres:var1}).
It is also worth noting that for both types of reservoir connectivity the results are the same as long as $G$ is less than the the length of cycles in the reservoir connectivity matrices.

\begin{figure}[tb]
\centering
\includegraphics[width=1.0\linewidth]{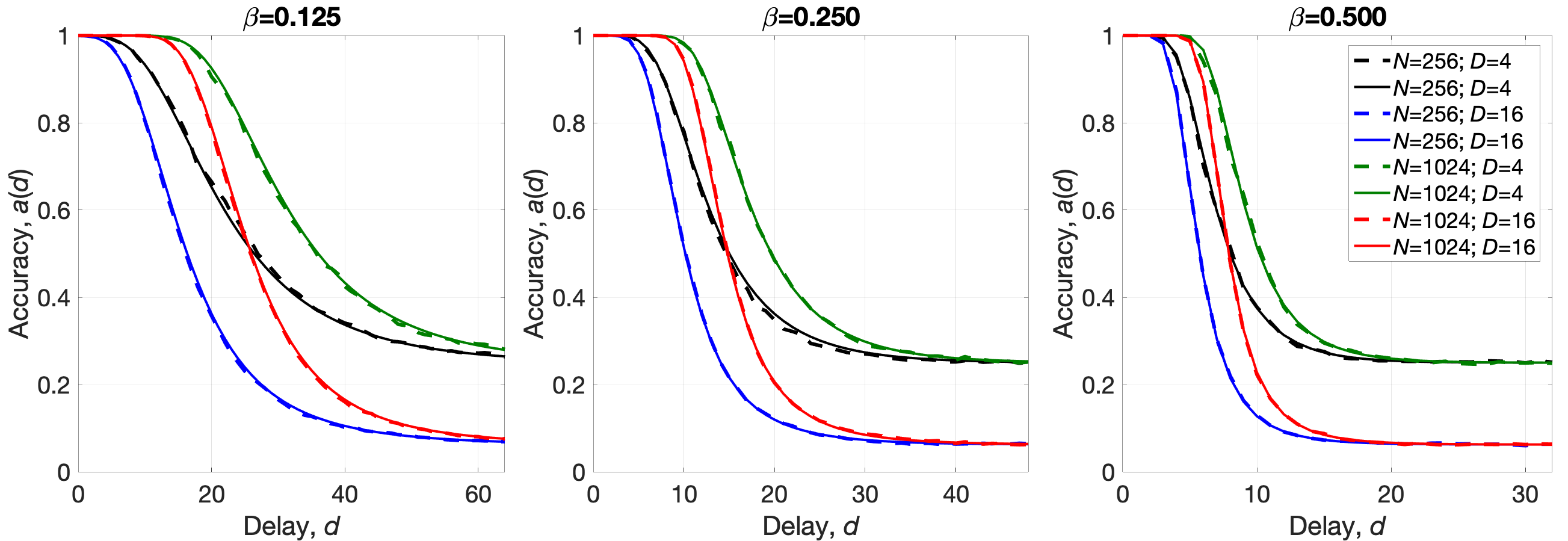}
\caption{
Predicted (solid lines) and experimental (dashed lines) recall accuracies for the reservoir with nonlinear transfer function and the input scaling  $\beta$, Eq.~(\ref{eq:esnres:var3}); $\mathbf{W}$ was a random permutation matrix; $\mathbf{W}^{\mathrm{out}}(d)$ was the codebook-based readout matrix.
Eq.~(\ref{eq:pcorr:orig1}) was used to obtain predicted accuracies.
$N$ was in $\{256, 1024\}$; $D$ was in $\{4, 16\}$.
The empirical results were averaged over $10$ simulation runs. Each simulation used a random sequence with $E=1,000$, $M=0$, and $R=3,000$. 
}
\label{fig:ESN:accuracy:tanh:beta}
\end{figure}

\paragraph{Linear reservoir with decay}

A recency effect can be included in a linear reservoir, Eq.~(\ref{eq:esnres:var1}), by adding the recurrent decay parameter $\gamma$, as in Eq.~(\ref{eq:esnres:var2}), and setting it to a value less than one. 
Fig.~\ref{fig:ESN:accuracy:gamma} shows predicted and experimental recall accuracies for Eq.~(\ref{eq:esnres:var2})  for three different values of $\gamma \in \{0.98, 0.90, 0.70\}$ and four combinations of $N \in \{256, 1024\}$ and $D \in \{4, 16\}$.
The recall accuracies in Fig.~\ref{fig:ESN:accuracy:gamma} are shown for different delay values. For larger $d$ values the recall accuracy $p_c(d)$ decreases until it finally reaches the level of the random guess accuracy ($1/D$).
Reducing the value of $\gamma$ steepens the decay of the accuracy curves, reaching the random guessing level more quickly.

\paragraph{Reservoir with nonlinear transfer function}

Another way to introduce the recency effect into the reservoir without $\gamma$ is to limit the values of $\mathbf{x}(n)$ by adding a nonlinear saturating transfer function~\cite{FradyNeCo2018}.
A common choice is $\tanh()$, which saturates at $-1$ and $+1$ for large negative and large positive arguments, respectively. 
In the corresponding dynamic equation, Eq.~(\ref{eq:esnres:var3}), the value of $\beta$ controls the recency effect. 
Fig.~\ref{fig:ESN:accuracy:tanh:beta} compares predicted and experimental recall accuracies for Eq.~(\ref{eq:esnres:var3}) as a function of delay for three different values of $\beta \in \{0.125, 0.250, 0.500\}$, four combinations of $N \in \{256, 1024\}$, and $D \in \{4, 16\}$.
Here larger values of $\beta$ impose stronger recency effects since $\beta$ controls the strength of the input to the reservoir relative to the recurrent activation.
Fig.~\ref{fig:ESN:accuracy:tanh:gamma:beta} in Supplementary Note~\ref{sec:tanh:decay} contains additional results for reservoir dynamics of Eq.~(\ref{eq:esnres:var4}) that combines the $\tanh()$ transfer function with a recurrent decay.

\subsubsection{Predicting ESN accuracies from measured statistical moments}
\label{sect:perf:measured}

\begin{figure}[tb]
\centering
\includegraphics[width=1.0\linewidth]{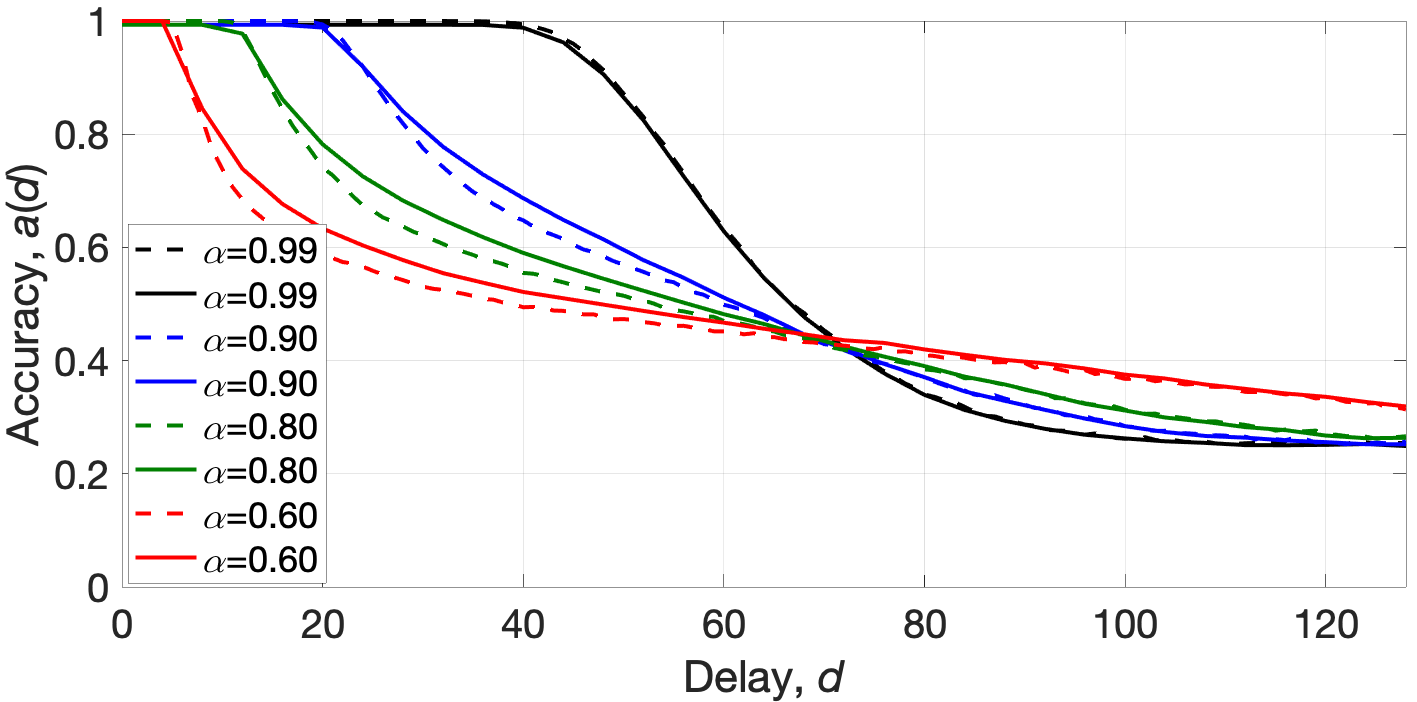}
\caption{
Predicted (solid lines) and experimental (dashed lines) recall accuracies for the reservoir updated according to Eq.~(\ref{eq:esnres:var5}); $\mathbf{W}$ was a random orthogonal matrix; $\mathbf{W}^{\mathrm{out}}(d)$ was the regression-based readout matrix.
Eq.~(\ref{eq:pcorr:mvn}) was used to obtain predicted accuracies.
$N$ was set to $256$; $D$ was to $4$.
The empirical results were averaged over $10$ simulation runs. Each simulation used random sequences with $E=1,000$, $M=8,192$, and $R=3,000$. 
}
\label{fig:ESN:accuracy:tanh:alpha:gamma:beta}
\end{figure}

Unfortunately, not all variants of ESN allow calculating the expected statistical moments of the reservoir directly from the model hyperparameters. 
Here, we show that the perceptron theory can quite decently predict even the performance of the most general variant of ESN, Eq.~(\ref{eq:esnres:var5}), when empirical measurements of the statistical moments of the readout network are available. 
Fig.~\ref{fig:ESN:accuracy:tanh:alpha:gamma:beta} depicts the predicted versus experimental recall accuracies for Eq.~(\ref{eq:esnres:var5}) as a function of the delay for four different values of $\alpha \in \{0.99, 0.90, 0.80, 0.60\}$ while the other hyperparameters were fixed: $N=256$; $D=4$; $\beta=0.0625$; $\gamma=0.98$.
We can see that $\alpha$ also controls recency effects, but in a more complicated way than the parameters $\gamma$ in Fig.~\ref{fig:ESN:accuracy:gamma} or $\beta$ in  Fig.~\ref{fig:ESN:accuracy:tanh:beta}. Large values of $\alpha$ produce a plateau of near optimal accuracy until to a drop-off at a threshold delay (for our parameter settings the drop off is around $d=40$). 
For smaller values of $\alpha$, the plateau of near optimal accuracy shortens, already falling off at smaller delays. 
However, there is a narrow inflection area at a specific delay where the curves for different $\alpha$ intersect. For delays larger than the inflection area (for our setting, around $d=70$), the reduction of $\alpha$ causes the inverse effect, a small increase in accuracy. Note that the theoretical predictions are precise for $\alpha$ near one, but slightly worse for smaller $\alpha$.

\subsubsection{ESN memory capacity in the full hyperparameter space}
\label{sect:perf:information}

As an aggregate single performance measure, we use the memory capacity of an ESN, that is, the maximum total information about the input sequence, $I_{\mathrm{tot}}$ measured in bits, that can be recalled at a point in the joint space of network's hyperparameters. Determining the capacity requires finding the optimal trade-off between sequence length and accuracy, for details see Supplementary Note~\ref{sec:information}. 
While a sweep of the entire joint parameter space with experiments would be computationally expensive, we use the perceptron theory to exhaustively investigate the parameter space of two different variants of ESN.

\begin{figure}[b]
\centering
\includegraphics[width=1.0\linewidth]{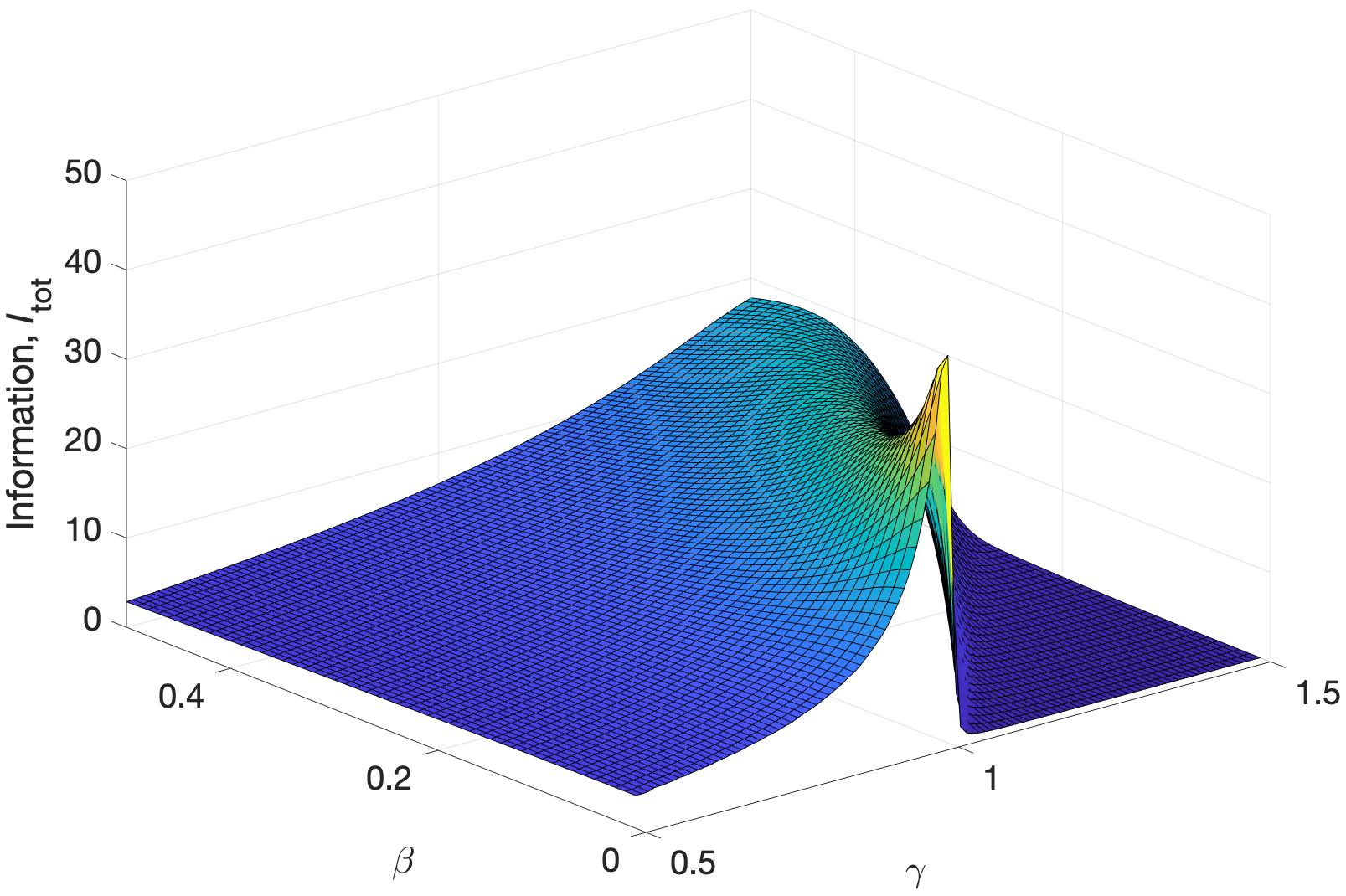}
\caption{
Memory capacity of ESN with dynamics according to Eq.~(\ref{eq:esnres:var4}).
The values were calculated analytically using Eq.~(\ref{eq:pcorr:orig1}).
$\mathbf{W}$ was a random permutation matrix; $\mathbf{W}^{\mathrm{out}}(d)$ was the codebook-based readout matrix; $N$ was set to $256$; $D$ was set to $2$;
$\beta$ varied in the range $[0.0, 0.5]$ with step of size $0.01$; 
$\gamma$ varied in the range $[0.5, 1.5]$ with step of size $0.01$.
}
\label{fig:ESN:information:tanh:gamma:beta}
\end{figure}

\begin{figure*}[tb]
\centering
\includegraphics[width=1.0\linewidth]{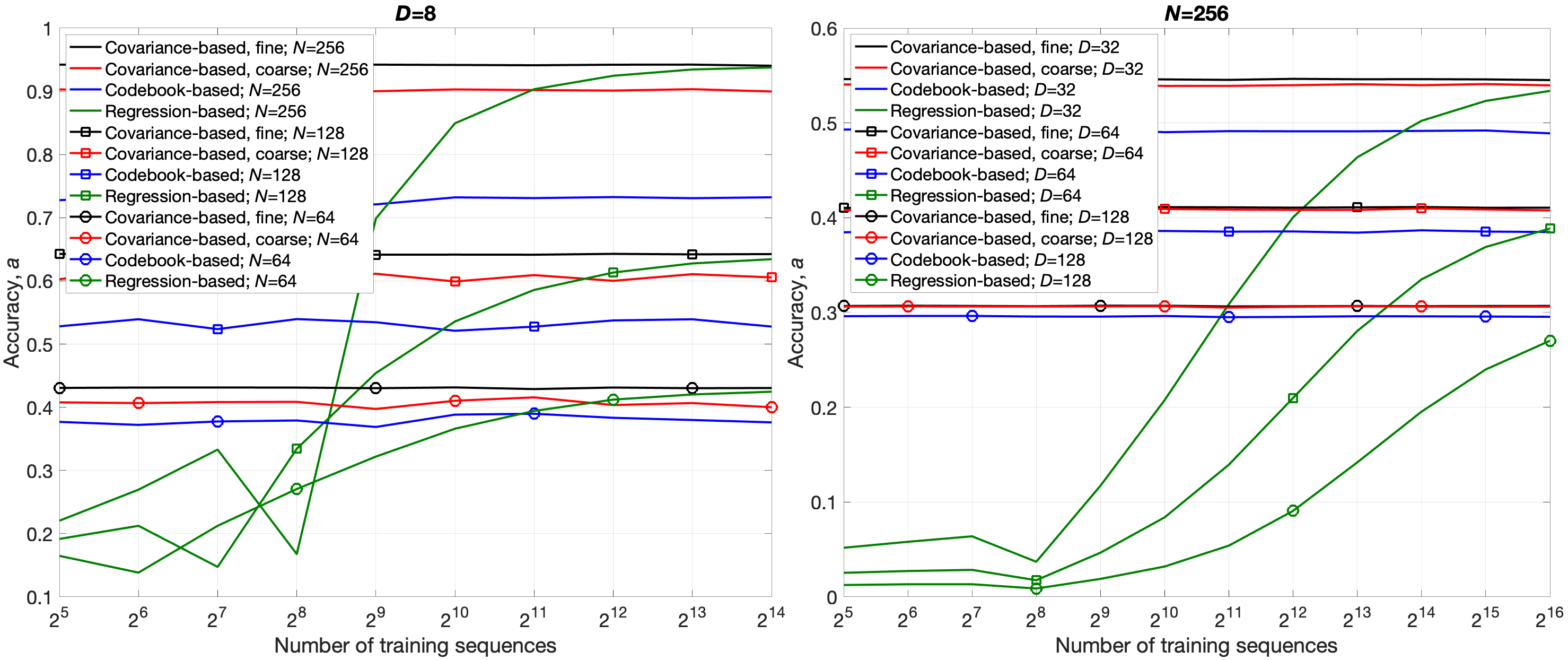}
\caption{
Experimental recall accuracy comparing the performance of the regression-based readout matrix for different number of training sequences to three other types of readout matrices, which do not require training. 
The reservoir dynamics is according to Eq.~(\ref{eq:esnres:var1}); $\mathbf{W}$ was a random orthogonal matrix (but results are the same for a random permutation matrix, not shown). 
Left panel: $D$ was fixed for three different values of $N$.
Right panel: $N$ was fixed for three different values of $D$.
In all experiments $G$ was fixed to $64$.
The results were averaged over $10$ simulation runs. 
Each simulation used $2,048$ random sequences to estimate the recall accuracy. 
}
\label{fig:ESN:accuracy:regression:data}
\end{figure*}


The first variant is the ESN with a reservoir dynamics evolving according to Eq.~(\ref{eq:esnres:var4}), $\mathbf{W}$ a random permutation matrix, and $\mathbf{W}^{\mathrm{out}}(d)$ the codebook-based readout matrix. For this model it is still possible to predict the accuracy of the network analytically (cf. Table~\ref{tab:ESN:theory}).
Fig.~\ref{fig:ESN:information:tanh:gamma:beta} shows the capacity $I_{\mathrm{tot}}$ when varying $\beta$ and $\gamma$.
As suggested by the previous experiments, the capacity decays quickly for increasing $\beta$ and decreasing $\gamma$. 
The largest capacity $I_{\mathrm{tot}}$ was observed for the smallest $\beta$ and $\gamma=1.00$. For larger settings of $\beta$, the highest information is reached at $\gamma \approx 1+\beta$. Thus, the optimal operation point is a recurrent matrix with a spectral radius of one and a very small input scaling coefficient. In this setting, the reservoir neurons stay for many iterations within the linear regime of their activation function, thereby delaying the erosion of stored information through nonlinear saturation effects. Although suboptimal because erosion sets in much earlier, for a larger input coefficient it is best to also proportionally adjust the spectral radius of the recurrent matrix to values larger than one.  

The second variant with a reservoir dynamics evolving according to Eq.~(\ref{eq:esnres:var4}) is reported in Supplementary Note~\ref{readout:inf:empir} (see Fig.~\ref{fig:ESN:information:alpha:tanh:gamma:beta}), illustrating that the capacity can be calculated even if the accuracy of the network cannot be predicted analytically.

\subsection{Effect of the readout type on ESN memory capacity}
\label{readouts:comparison}

The previous sections focused on models with codebook-based readout matrices since in this case the accuracy can be predicted analytically.  
However, as explained in Sections~\ref{sect:readout}  and~\ref{sect:restuls:covariance}, there are several other types of readout for the trajectory association task. 
Here, we will assess the effect of the readout matrix type for two variants of the reservoir dynamics: linear, Eq.~(\ref{eq:esnres:var1}), and with recurrent decay, Eq.~(\ref{eq:esnres:var2}). 

\begin{figure}[tb]
\centering
\includegraphics[width=1.0\linewidth]{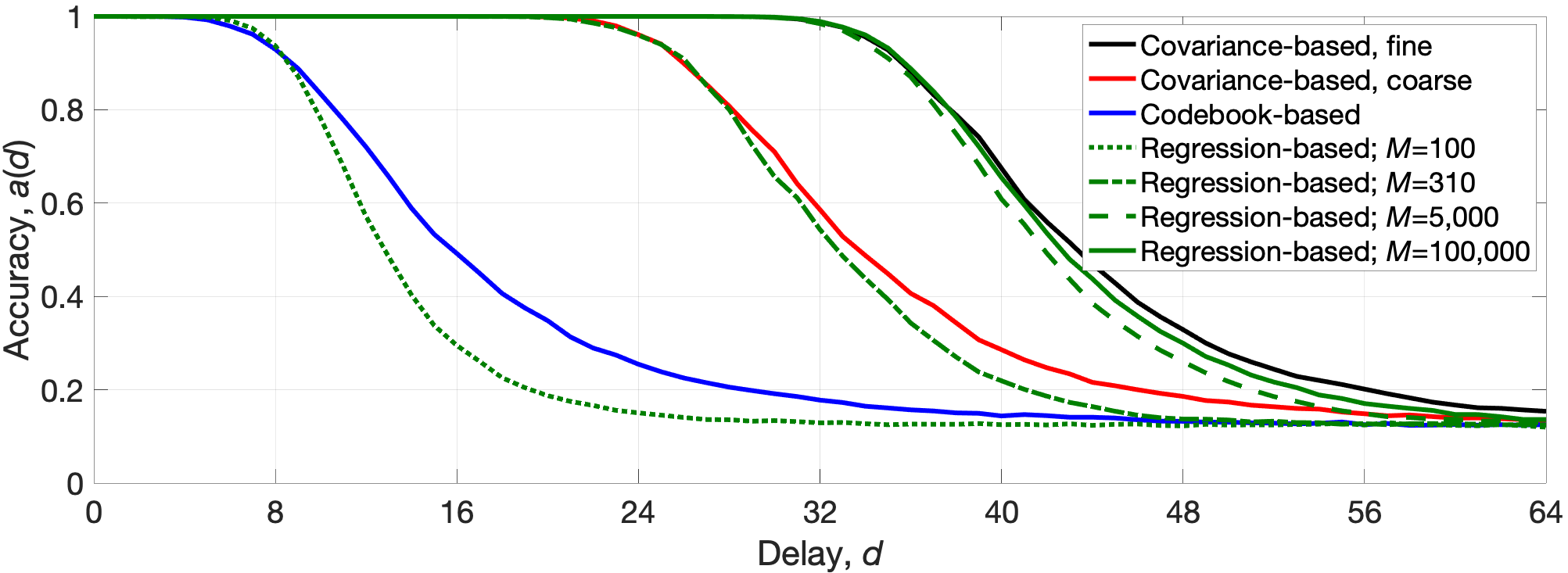}
\caption{
Experimental recall accuracy for different readout types for reservoir dynamics in Eq.~(\ref{eq:esnres:var2}); $\mathbf{W}$ was a random orthogonal matrix; $N$ was set to $256$; $D$ was set to $8$; $\gamma$ was set to $0.9$.
The results were averaged over $10$ simulation runs. 
For each readout type, each simulation used a random sequence with $E=1,000$, and $R=3,000$. 
For the regression-based readout, the value of $M$ is indicated in the legend.
For all other readout types $M=0$.
}
\label{fig:ESN:accuracy:readouts:linear}
\end{figure}

\begin{figure*}[tb]
\centering
\includegraphics[width=1.0\linewidth]{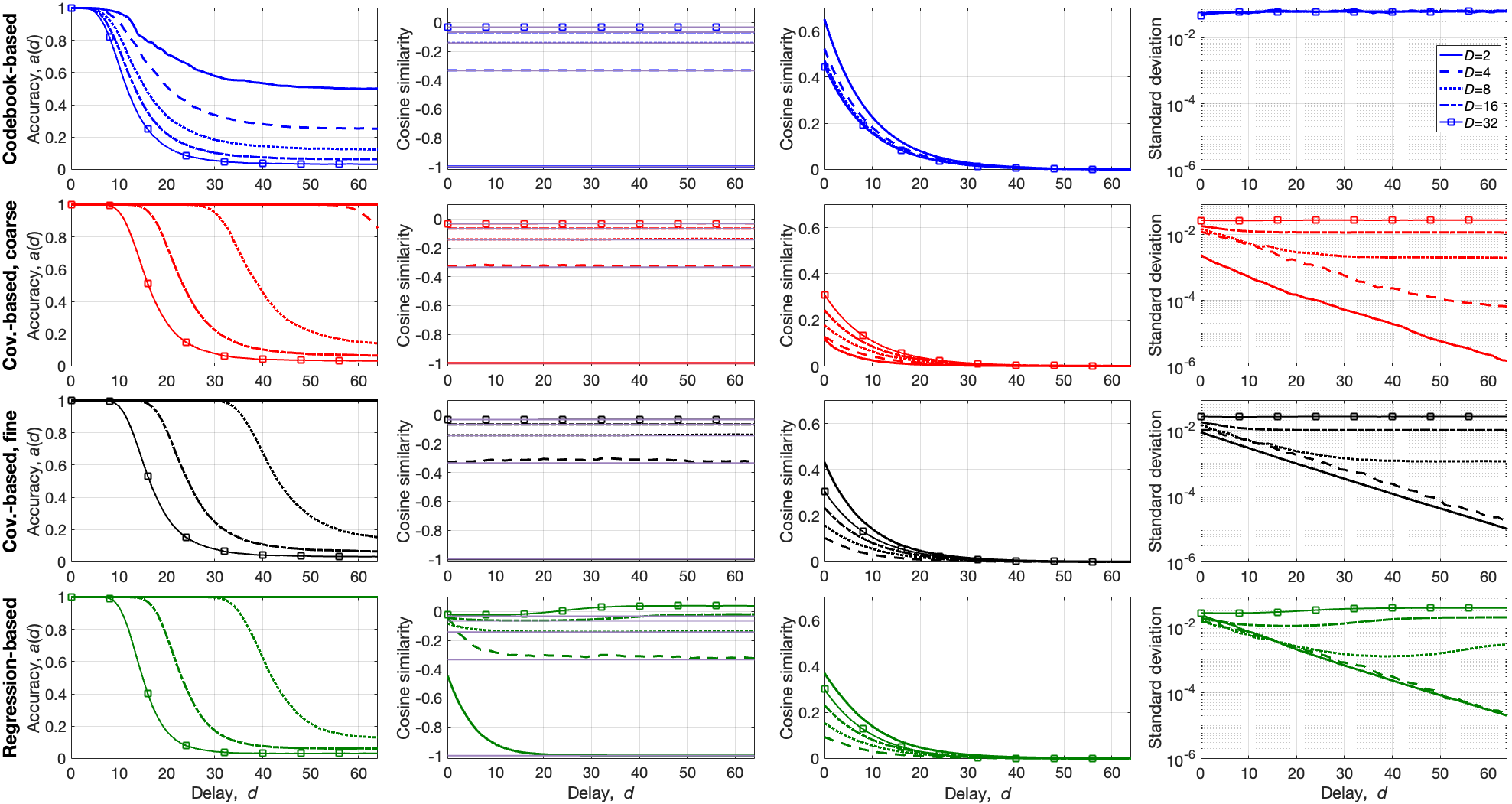}
\caption{
Experiments exploring the geometric structure of readout matrices of different types for reservoir dynamics in Eq.~(\ref{eq:esnres:var2}).
All measured values were plotted against the delay $d$.
$N$ was set to $256$; $D$ varied between $\{2, 4, 8, 16, 32 \}$ (see legend), $\gamma$ was set to $0.9$.
Each row corresponds to a particular type of the readout matrix. 
The first column reports the recall accuracies. 
The second column reports the cosine similarity between the entries of the readout matrix.
The third column reports the difference between $\mu_{h}$ and $\mu_{r}$ but for the ease of comparison they were measured as cosine similarities and not as inner products.
The fourth column reports $\sigma_{h}$ on the logarithmic scale.
The results were averaged over $20$ simulation runs, each simulation used a random sequence with $E=1,000$, $M=5,000$, and $R=5,000$. 
}
\label{fig:ESN:readouts:simplex}
\end{figure*}

\subsubsection{Linear update of the reservoir}
\label{readouts:linear}

Fig.~\ref{fig:ESN:accuracy:regression:data} shows experimental recall accuracies for the reservoir updated according to Eq.~(\ref{eq:esnres:var1}) for four different types of readout matrices against the number of training sequences. 
Note that only the regression-based readout matrices use the training data so the accuracy of other approaches are, essentially, horizontal lines.\footnote{Methods without training show small deviations from flat horizontal lines caused by random initialization of codebooks and sequences during individual simulation runs.} 
Left panel in Fig.~\ref{fig:ESN:accuracy:regression:data}  depicts the results when the value of $D$ was fixed to $8$; the value of $N$ was in $\{ 64, 128, 256 \}$.
Right panel depicts the results when the value of $N$ was fixed to $256$; the value of $D$ was in $\{ 32, 64, 128 \}$.

For both settings, we clearly see that among readout matrices not relying on training data and independent of $N$ and $D$, the codebook-based readout matrix exhibited the lowest performance, followed by the coarse covariance-based readout matrix, while the novel fine-grained covariance-based readout matrix performed best. 
The actual difference in performance between these readout matrices depends also on $N$ and $D$. For example, when $N=256$ \& $D=8$ (left panel) the difference between the codebook-based readout matrix and covariance-based readout matrices is big, while for $N=256$ \& $D=128$ (right panel) the difference was rather small. 
In general, we observe that for fixed $D$, the performance difference increases as $N$ grows. Conversely, for $N$ fixed, the performance difference decreases as $D$ grows.

As expected, we observe that the accuracy of regression-based readout matrices depends on the amount of available training data. With sufficient data, their accuracy approaches that of the novel fine-grained covariance-based readout matrix. 
When $D$ is fixed (left panel), the amount of data required to match the accuracy of the fine-grained covariance-based readout matrix is roughly the same across different $N$. However, for larger $N$, the accuracy curves are steeper, indicating that less data is needed to reach the performance of the corresponding codebook-based and coarse covariance-based readout matrices. In contrast, when $N$ is fixed (right panel), larger $D$ requires more data to approach the accuracy of the fine-grained covariance-based readout matrix. For example, with $2^{16}$ training sequences, at $D=32$ the accuracy of the regression-based readout matrix nearly matches that of the coarse covariance-based readout matrix;  at $D=64$ it matches the codebook-based readout matrix; while at $D=128$ it falls short even of the codebook-based readout matrix.

\subsubsection{Reservoir with recurrent decay}
\label{readouts:decay}

According to Table~\ref{tab:ESN:theory}, the covariance-based regression matrices can still be calculated with Eqs.~(\ref{eq:cov:coarse:gamma}) and (\ref{eq:cov:fine:gamma}) when the recurrent decay $\gamma$ is used during the reservoir update. 
Fig.~\ref{fig:ESN:accuracy:readouts:linear} shows the experimental recall accuracies for the reservoir with a dynamics of Eq.~(\ref{eq:esnres:var2}) for four different types of readout matrices. 
The hyperparameters of the network were set as follows: $N=256$, $D=8$, $\gamma=0.9$.
Unlike for the linear reservoir dynamics, Eq.~(\ref{eq:esnres:var1}), the accuracy depends on the setting of $d$, displayed on the $x$-axis. 
We find that the performance of readout types without training is ranked in the same order as before in Fig.~\ref{fig:ESN:accuracy:regression:data}: codebook-based on the low end, coarse covariance-based and the novel fine-grained covariance-based readout as the best. 
For each of these readout types, there is a corresponding curve for the regression-based readout matrix where the size of the training data was chosen such that the high fidelity mode of the regression-based readout matrix matches that of the corresponding baseline. 
Note that near the curve for the fine-grained covariance-based readout matrix, there are two curves for the regression-based readout matrix.
These curves demonstrate the diminishing returns of additional training data when the performance is already close to that of the fine-grained covariance-based readout matrix.
In particular, when the length of the memorization phase increased $20$ times from $5{,}000$ to $100{,}000$, the improvement in the recall accuracy was barely noticeable.

\subsection{Geometric structure of the readout matrices} 
\label{sec:results:readouts:structure}
The recent study~\cite{papyan2020prevalence} has shown that in deep neural networks during the terminal phase, the last layer of the network (i.e., the readout network) approaches the special geometric structure -- equiangular tight frame that contains regular simplices (simplex ETF). 
The entries of the simplex ETF (i.e., the rows of $\mathbf{W}^{\mathrm{out}}(d)$) have the Euclidean norms equal to one and the cosine similarity between different entries equals $-\frac{1}{D-1}$. 
Here, we investigate whether the readout matrices in different ESNs models exhibit a similar simplex ETF structure.
The experiments were done for all four different types of readout networks, shown in the four rows of panels in Fig.~\ref{fig:ESN:readouts:simplex}.
The amount of training data (i.e., $M$) was chosen such that the regression-based readout would have an accuracy similar to the fine-grained covariance-based readout matrix. The accuracy curves in the first column of Fig.~\ref{fig:ESN:readouts:simplex} are consistent with the results in Sections~\ref{readouts:linear} and~\ref{readouts:decay}. 

The second column in Fig.~\ref{fig:ESN:readouts:simplex} presents the cosine similarities between the rows of the readout matrices. For visual guidance, the thin solid horizontal lines represent the cosine similarities predicted by the simplex ETF structure for different $D$. Interestingly, the cosine similarities of codebook-based and both covariance-based readout matrices coincide with the simplex ETF structure for all values of $D$.

The results for regression-based readout is more nuanced.  For small values of $D \in \{2,4\}$, 
the training converged to the simplex ETF structure for larger $d$.
For larger values of $D$, the simplex ETF structure was present in the beginning but the training moved the cosine similarities gradually away from the ETF structure. 
We hypothesize that both of these observations are related to the strength of symbols' traces in the reservoir state: when the signal is very strong, there is no need to converge to the simplex ETF structure to detect it. 
Conversely, when the signal is too weak, there is not enough information to recover the simplex ETF structure.
This hypothesis was tested and confirmed in Supplementary Note~\ref{readout:reg:noise}.

The last two columns show how the difference between $\mu_{h}$ and $\mu_{r}$ as well as $\sigma_{h}$ changed for different values of $d$ and $D$.\footnote{
value of $\sigma_{r}$ were similar to $\sigma_{h}$ and so they were omitted.
}
This behavior provides insights into why different readout matrices demonstrated different recall accuracies. 
The perceptron theory suggests that the statistical moments of inner products are crucial in predicting the expected accuracy of the recall.
What is less intuitive, however, is that different types of readout matrices identified different compromises between $\mu_{h}$, $\mu_{r}$, $\sigma_{r}$, and $\sigma_{h}$.
The codebook-based readout matrix is the most intuitive one where the $\sigma_{h}$ is the same for different values of $D$ and $d$.
Other types (especially for smaller $D$) allowed smaller values of $\mu_{h}$ at the cost of significantly decreasing $\sigma_{r}$ and $\sigma_{h}$.
A notable example is the coarse covariance-based readout matrix for $D=2$, where $\mu_{h}$ was significantly smaller than that of the codebook-based readout matrix but at the same time  $\sigma_{r}$ was at least an order of magnitude smaller. 
Such choice of the statistical moments allowed for the perfect recall in the considered range of $d$ while the accuracy of the codebook-based readout matrix started to decrease already after $d>15$.
It is worth noting, however, that for larger values of $D$, the behaviors of the other readout matrices is becoming more similar to the behavior of the codebook-based readout matrix. 
For example, for $D=32$ the values of $\sigma_{r}$ were the largest ones and they were not changing much with $d$; $\mu_{h}-\mu_{r}$ curves  were also getting similar to the corresponding curve for the codebook-based readout matrix.
As a result, the recall accuracy curves for different readout matrices  when $D=32$ are very similar to each other.
Supplementary Note~\ref{sect:suppl:zscore} augments these qualitative observations using $Z$-scores to elucidate compromises observed for different types of readout matrices in a single metric and leads to additional quantitative observations.

\section{Discussion}
\label{sect:discussion}

\subsection{Summary of results} 

Echo state networks (ESN) are artificial neural networks that consist of a recurrent neural network with fixed weights, followed by a perceptron-like readout network. Over time, the ESN literature has accumulated a plethora of models differing in their recurrent dynamics and how the perceptron weights are trained or computed.  Here, we recapitulated various approximation levels of perceptron theory (Section~\ref{sec:capacity} and Supplementary Note~\ref{sec:theory:comparisons}) and investigated their ability to predict the performance of $30$ different types of ESNs and their working memory capacities.

The studied ESN models range from very simple (e.g., permutation-based reservoir connectivity, linear reservoir updates, and codebook-based readout) to more complex (e.g., leaky integration, reservoir connectivity with random orthogonal matrix, $\tanh()$ transfer function, and regression-based readout). Inspired by perception theory, in Section~\ref{sect:restuls:covariance} we extended the repertoire of ESN models, by proposing two novel readout networks where the weights can be precomputed even if the reservoir states are correlated.

Comparing theoretical predictions to extensive experiments, we found that perceptron theory can analytically predict the memory capacity of many variants based on the networks' hyperparameters (Table~\ref{tab:ESN:theory}). Further, any model can be analyzed with the perceptron theory, if the first two moments (mean and covariance) of the inner-product statistics are measured.
These results have practical implications.  The ability to predict memory capacity enables the principled design of linear ESNs that store any desired amount of information about the inputs. The reservoir state can then also be passed to another recurrent network forming higher-order polynomial features~\cite{kleyko2025principled}, closely approximating the recently proposed nonlinear vector autoregression formulation of reservoir computing~\cite{gauthier2021next}.

To assess the joint influence of hyperparameters on working memory capacity, we used the perceptron theory to estimate capacity values over the entire parameter space (Supplementary Note~\ref{sec:information}). We found that the capacity peaks when the recurrent decay $\gamma$ is close to one and the input scaling  $\beta$ is close to zero (cf. Fig.~\ref{fig:ESN:information:tanh:gamma:beta}).

We also compared the theory to ESNs with different types of precomputed readout networks (Sections~\ref{sect:readout}), including the novel models (Section~\ref{sect:restuls:covariance}). 
For two different reservoir dynamics, linear and with recurrent decay, these models exhibited performance levels that are commensurate with the computational complexity of precomputing the output weights (Figs.~\ref{fig:ESN:accuracy:regression:data} and~\ref{fig:ESN:accuracy:readouts:linear}). 
Specifically, the novel ESN model with fine-grained covariance-based readout performed significantly better than the simpler models.
However, this performance gap narrowed as the alphabet size increased or the reservoir size decreased. 
Regression-based readout networks can achieve the performance of the fine-grained covariance-based readout, given sufficient training data. Thus, the novel ESN model is particularly interesting in cases when training data are scarce.

Next, we show that the geometry of the ESN readout networks can be analyzed in a similar way as the last layer of a deep neural network. 
It has been reported that for deep networks, the training forces the rows in the weight matrix into 
equiangular tight frames containing regular simplices, a phenomenon referred to as ``neural collapse''~\cite{papyan2020prevalence}. 
We found that many ESN readouts also exhibit neural collapse: specifically, all the models with precomputed readout networks, as well as regression-based readout networks for many parameter settings. 
We also demonstrated that the difference in the readout types can be characterized in terms of trade-off between the statistical moments of inner products (see Supplementary Note~\ref{sect:suppl:zscore}): either a high mean with a high standard deviation, or a much smaller standard deviation at the cost of a reduced mean. 
We further showed that differences among readout types can be characterized as a trade-off in the statistical moments of inner products (Supplementary Note~\ref{sect:suppl:zscore}), with noise directly influencing this trade-off (Supplementary Note~\ref{readout:reg:noise}).

Finally, Supplementary Note~\ref{sect:perf:5:bit} demonstrated how the perceptron theory can be used to analytically predict the reservoir size required to solve $5$-Bit Memory Task (Section~\ref{sect:methods:trajectory}), a widely used benchmark for recurrent neural networks~\cite{JaegerMemoryESN2012}.
The predicted reservoir sizes closely matched empirical averages across different distractor periods.
To our knowledge, no other theory provides such predictions for this task.

\subsection{Related work and connections to this study} 
\label{sec:related}


Our experiments with various ESN models connect to earlier investigations. The memory capacity, the channel capacity between inputs and readout of the ESN, was first introduced as a performance measure in~\cite{JaegerMemory2002}, and experimentally evaluated for comparing ESN variants, e.g.~\cite{wringe2025reservoir}, and understand influences of external noise and internal crosstalk.      
These studies revealed that ESNs with linear reservoir can exhibit perfect accuracy for sequences shorter than the reservoir dimension, but deteriorating accuracy for delays longer than the reservoir dimension~\cite{Fette2005ShortSimple, MaMemoryESN2014} (cf. Fig.~\ref{fig:ESN:accuracy:linear}).  
In~\cite{JaegerMemory2002}, it was shown that the memory capacity of a reservoir quantified as correlations between true inputs and their linear reconstructions from reservoir states is bounded by the reservoir dimension, assuming i.i.d. inputs and linear readout. 

For linear reservoirs, analytical capacity curves were derived under orthogonal connectivity and recurrent decay (cf. Eq.~(\ref{eq:esnres:var2}))~\cite{WhiteOrthogonal2004}, closely matching our theoretical predictions in Fig.~\ref{fig:ESN:accuracy:gamma}. The influence of input sparsity on linear ESNs was quantified in~\cite{GanguliShortTerm2010, CharlesShortTerm2014}, demonstrating that sparse input leads to superlinear scaling of capacity with reservoir size.

Further, the above studies, and some other experimental studies~\cite{VerstraetenNonlinearityESN2010, GoudarziNonlinearityESN2015}, investigated ESNs with nonlinear reservoir neurons. In these models, increasing $\beta$ or $\gamma$ shortens the memory window resulting in a memory buffer function, in which older inputs are gradually forgotten (cf. Fig.~\ref{fig:ESN:accuracy:tanh:beta} and cf. Supplementary Note~\ref{sec:tanh:decay}).
Extensions of analytical investigations to nonlinear reservoirs remain scarce; notable examples include~\cite{HarunaOptimalShortTerm2019, GononMemoryForecasting2020, takasu2025neuronal}. In~\cite{HarunaOptimalShortTerm2019}, mean-field theory~\cite{MassarMeanFieldESN2013} was applied to ESNs with $\erf()$ transfer functions, focusing on small $\beta$ values that render the dynamics nearly linear. Capacity scaling with correlated inputs or reservoir states were studied in~\cite{GononMemoryForecasting2020} and~\cite{takasu2025neuronal}, respectively. In~\cite{takasu2025neuronal}, it was shown that sublinear scaling of memory capacity is predicted by the strength of correlations in neuronal activity, and this phenomenon can be observed across a variety of recurrent neural networks.

Other experimental studies~\cite{FarkavsComputational2016, Gallicchio2020Sparsity} examined the effect of sparsity of the matrices in ESNs with nonlinear neurons. They found that sparser input projections $\mathbf{W}^{\mathrm{in}}$ lead to higher capacity. By contrast, sparsity in the recurrent matrix $\mathbf{W}$ was found to have little effect on memory capacity, consistent with our results of finding high capacity in ESNs with the recurrent weights being a (very sparse) permutation matrix.

The study of optimal reservoir design in ESNs was initiated in~\cite{OzturkAnalysis2007}, which compared memory capacity under different initialization strategies for the connectivity matrix. A key finding was that matrices with the same spectral radius can exhibit markedly different performance depending on the initialization design choice. In~\cite{RodanMinESN2011, intESN}, a particularly simple connectivity -- circular shift matrix -- was investigated. This matrix, a special case of the permutation matrices considered here, offered a minimal design. Building on this direction,~\cite{StraussDesign2012} proposed several deterministic methods for constructing sparse orthogonal connectivity matrices with controllable density. These designs, which included circular shift matrix, were shown to yield higher memory capacity than random reservoir connectivity matrices.

\label{sect:bib}
\bibliographystyle{unsrt}
\bibliography{references}

\newpage

\phantom{a}
\newpage


\setcounter{equation}{0}
\setcounter{figure}{0}
\setcounter{table}{0}
\setcounter{section}{0}
\setcounter{page}{1}
\makeatletter
\renewcommand{\thesection}{S-\Roman{section}}
\renewcommand{\theequation}{S.\arabic{equation}}
\renewcommand{\thefigure}{S.\arabic{figure}}
\renewcommand{\thetable}{S.\arabic{table}}

\begin{strip}
\noindent
\textbf{\LARGE Supplementary Material: Towards a Comprehensive Theory of \\Reservoir Computing} 
\end{strip}

\section{Quantifying information retrieved from the reservoir}
\label{sec:information}

The performance of an echo state network across the entire range of recall delays can be assessed by the total information that can be stored and retrieved from the reservoir.
Following~\cite{FradyNeCo2018}, the amount of information retrieved at a certain delay $d$ can be estimated (for details, see Section 2.2.3) as:  
\noindent
\begin{equation}
\label{eq:MI:item}
\begin{split}
I(p_c(d)) = & p_c(d) \log_2(D p_c(d) ) + \\
& +  (1-p_c(d)) \log_2 \left( \frac{D}{D-1} (1-p_c(d)) \right).
\end{split}
\end{equation}
\noindent
Note that the retrieved information according to Eq.~(\ref{eq:MI:item}) is zero if the accuracy is equal to random guessing ($p_c(d) = 1/D$).
The total amount of information retrieved from the reservoir is calculated as the sum of information extracted for individual delays: 
\noindent
\begin{equation}
\label{eq:MI:tot}
I_{\mathrm{tot}}=\sum_{d=0}^{\infty} I(p_c(d)).
\end{equation}
\noindent
In practice, the infinite sum can be truncated at a value of $d$ where $p_c(d)$ approximates $1/D$ and the amount of information becomes negligible.

\begin{figure}[tb]
\centering
\includegraphics[width=1.0\linewidth]{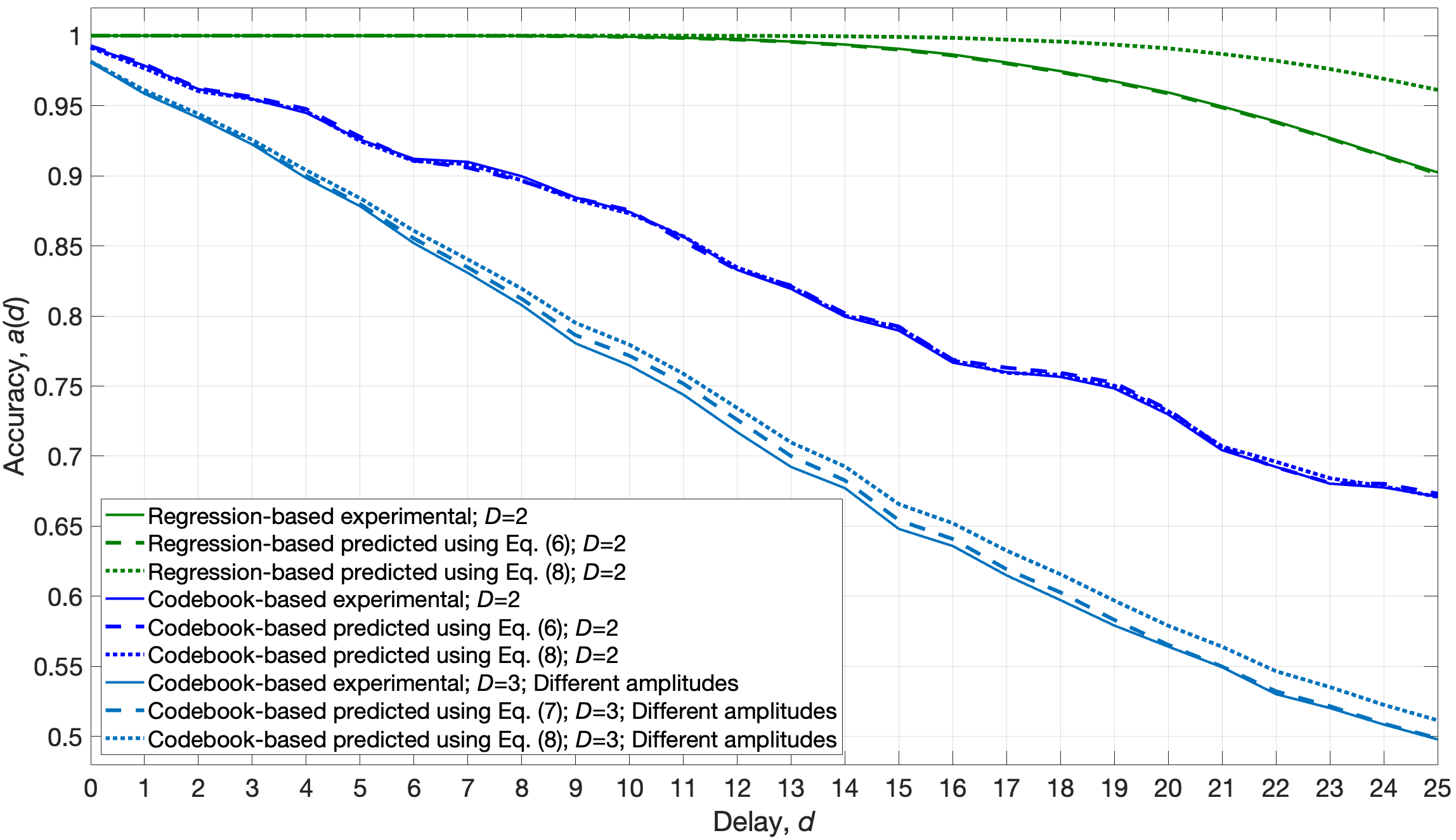}
\caption{Experimental (solid lines) and predicted (dashed and dotted lines) recall accuracies for different approximation levels of the perceptron theory. 
$\mathbf{W}$ was a random permutation matrix; the type of $\mathbf{W}^{\mathrm{out}}$ is specified in the legend for each line; $N=100$, $D=2$ (for red and blues lines) or $D=3$ (for cyan lines), $\kappa = 4$. The empirical results were averaged over $50$ simulation runs. The experiments are reproduced from~\cite{KleykoPerceptron2020}.  
}
\label{fig:ESN:acc:perc}
\end{figure}

\section{Predicting performance with different approximation levels of the perceptron theory}
\label{sec:theory:comparisons}

To compare different approximation levels of the perceptron theory, Eqs.~(\ref{eq:pcorr:mvn})-(\ref{eq:pcorr:orig1}), we replicate an experiment from~\cite{KleykoPerceptron2020}. Fig.~\ref{fig:ESN:acc:perc} presents the average experimental recall accuracies for two echo state network models with different readout networks: the regression-based readout (green solid line) and the codebook-based readout (blue solid line).

The dotted lines in Fig.~\ref{fig:ESN:acc:perc} show theoretical predictions derived from measured statistics of inner products. The approximation given by Eq.~(\ref{eq:pcorr:orig1}) accurately captures the performance of the codebook-based readout but overestimates that of the regression-based readout, as it neglects the correlation between the rows of the regression-based readout matrix. Fig.~\ref{fig:ESN:acc:perc} also depicts predictions obtained using Eq.~(\ref{eq:pcorr:mvn}) (dashed lines) that correctly describe the accuracy of both models. Predictions based on Eq.~(\ref{eq:pcorr:indep}) are omitted since, for $D=2$, they coincide with those from Eq.~(\ref{eq:pcorr:orig1}).

To highlight the differences between these approximations, an additional experiment was performed with $D=3$ using the codebook-based readout and inputs of varying amplitudes (cyan solid line). In this case, Eq.~(\ref{eq:pcorr:orig1}) (cyan dotted line) overestimates the accuracy, whereas Eq.~(\ref{eq:pcorr:indep}) (cyan dashed line) provides an accurate prediction.

\begin{figure}[t]
\centering
\includegraphics[width=1.0\linewidth]{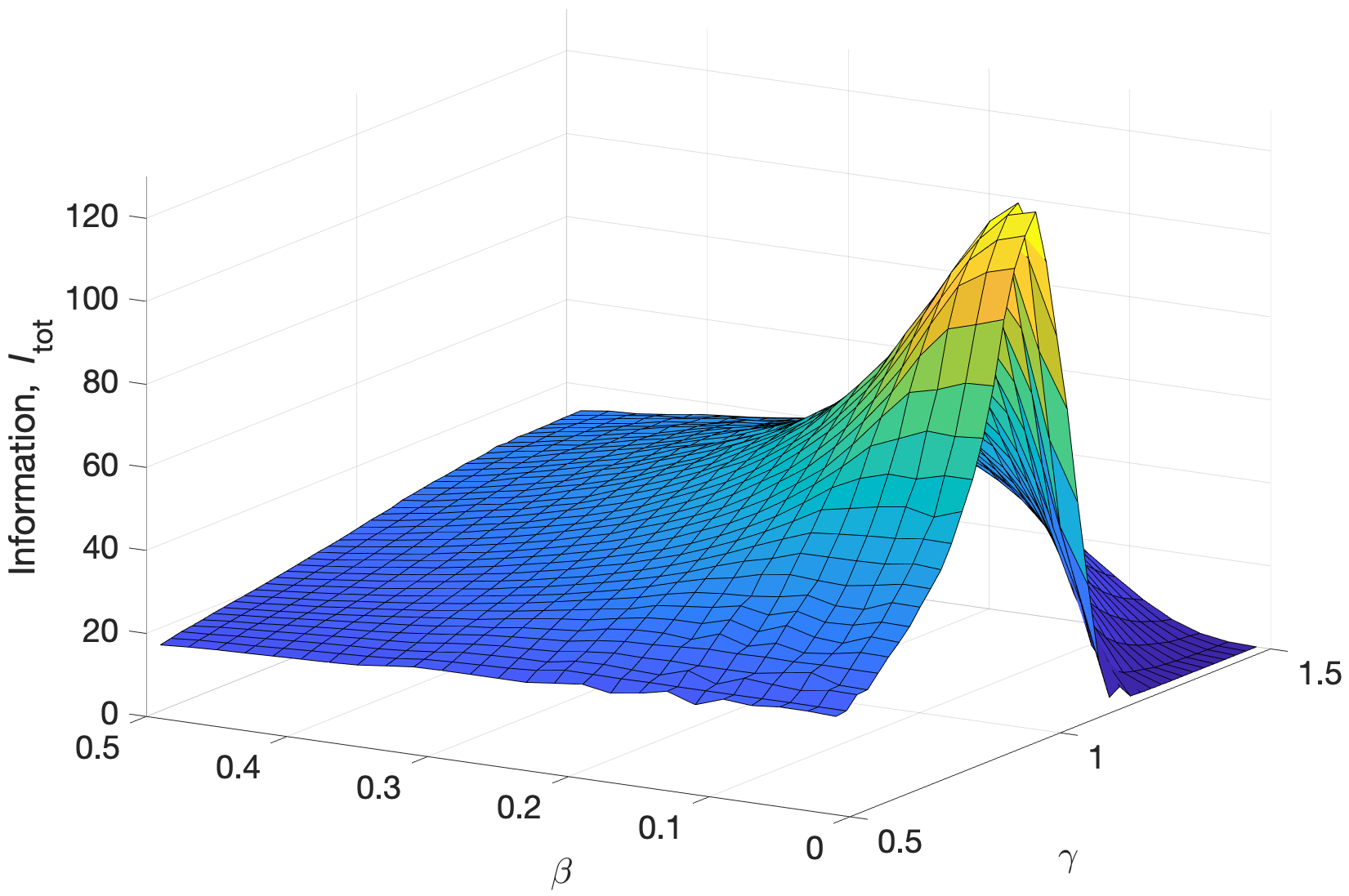}
\caption{
The amount of information extracted from the reservoir updated according to Eq.~(\ref{eq:esnres});
The values of $I_{\mathrm{tot}}$ were calculated from empirically estimated $p_c(d)$ when $d$ was in the range $[0,2N]$.
$\mathbf{W}$ was a random orthogonal matrix; $\mathbf{W}^{\mathrm{out}}(d)$ was the regression-based readout matrix.
$N$ was set to $256$; $D$ was $8$; $\alpha$ was $0.99$;
$\beta$ varied in the range $[0.0, 0.5]$ with step of size $0.01$; 
$\gamma$ varied in the range $[0.5, 1.5]$ with step of size $0.01$.
The empirical values of $p_c(d)$ were averaged over $10$ simulation runs. 
Each simulation used a random sequence with $E=1,000$, $M=4,096$, and $R=3,000$. 
}
\label{fig:ESN:information:alpha:tanh:gamma:beta}
\end{figure}

\section{Effect of hyperparameters on the amount of information recalled from the reservoir for empirically measured accuracy}
\label{readout:inf:empir}

Note that $I_{\mathrm{tot}}$ can be calculated even if the accuracy of the network cannot be predicted analytically. 
For example, we can do so for the case when the reservoir is updated according to Eq.~(\ref{eq:esnres}) and $\mathbf{W}^{\mathrm{out}}(d)$ is the regression-based readout matrix.
Fig.~\ref{fig:ESN:information:alpha:tanh:gamma:beta} presents $I_{\mathrm{tot}}$ for varying values of $\beta$ and $\gamma$. 
The other hyperparameters were set as follows: $N=256$; $D=8$; $\alpha=0.99$.
The observed values differed from the ones in Fig.~\ref{fig:ESN:information:tanh:gamma:beta} but qualitatively the behavior of the surface was the same.

\begin{figure*}[t]
\centering
\includegraphics[width=0.95\linewidth]{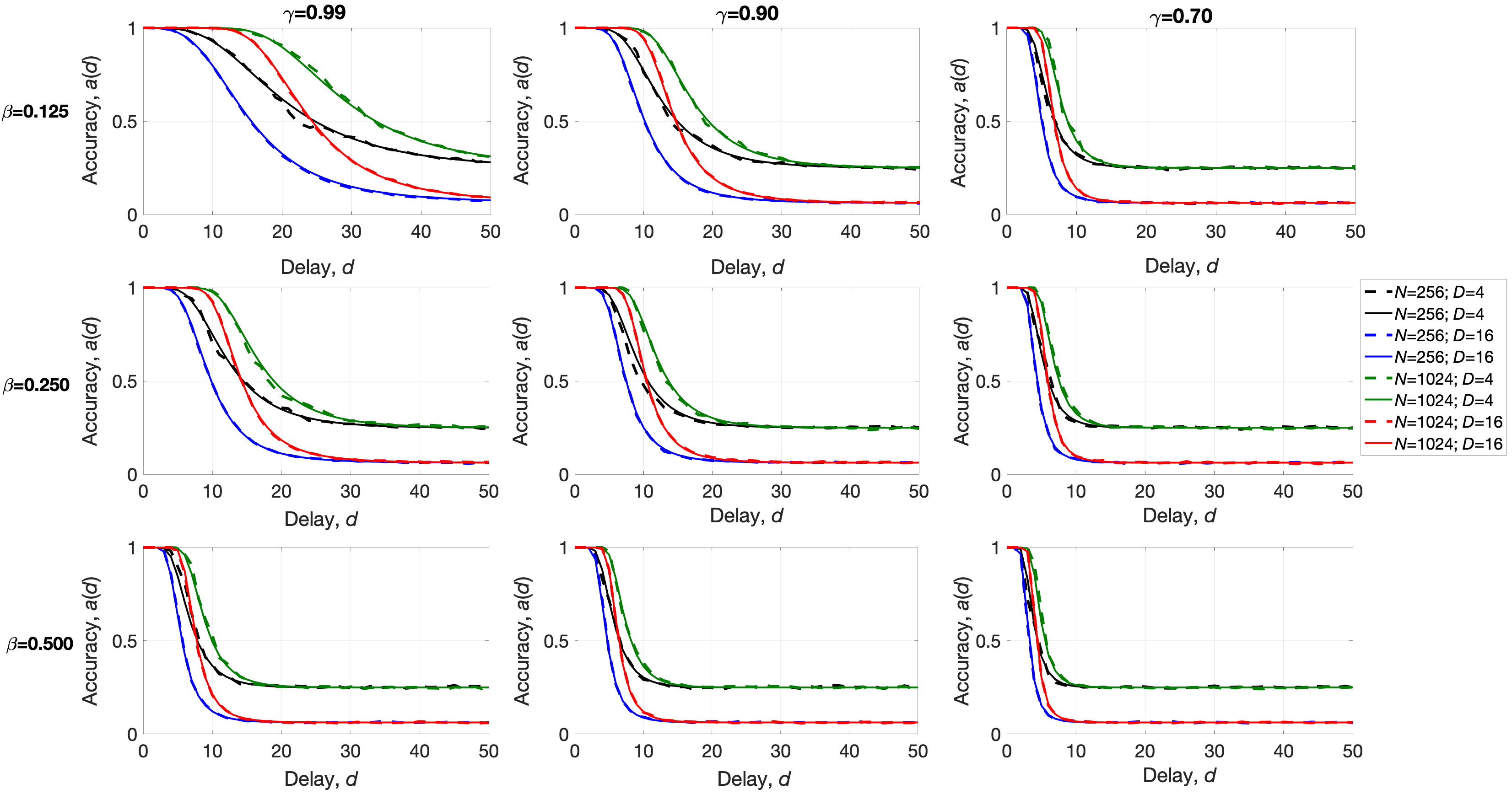}
\caption{
Accuracy of the recall for analytical results (solid lines) versus experimental results (dashed lines) for the reservoir updated according to Eq.~(\ref{eq:esnres:var4}); $\mathbf{W}$ was a random permutation matrix; $\mathbf{W}^{\mathrm{out}}(d)$ was the codebook-based readout matrix.
Eq.~(\ref{eq:pcorr:orig1}) was used to obtain predicted accuracies.
$N$ was in $\{256, 1024\}$; $D$ was in $\{4, 16\}$.
The empirical results were averaged over $10$ simulation runs. Each simulation used a random sequence with $E=1,000$, $M=0$, and $R=3,000$. 
}
\label{fig:ESN:accuracy:tanh:gamma:beta}
\end{figure*}

\section{Reservoir with nonlinear transfer function and recurrent decay}
\label{sec:tanh:decay}

Despite both reservoir updates Eq.~(\ref{eq:esnres:var2}) and Eq.~(\ref{eq:esnres:var3}) control the recency effect individually, it is common to combine them together as in Eq.~(\ref{eq:esnres:var4}). 
Then a combination of $\beta$ and $\gamma$ will determine the accuracy curve.   
Fig.~\ref{fig:ESN:accuracy:tanh:gamma:beta} presents both analytical and empirical recall accuracies for Eq.~(\ref{eq:esnres:var4}) against different $d$ for nine different combinations of $\beta \in \{0.125, 0.250, 0.500\}$ and $\gamma \in \{0.99, 0.90, 0.70\}$ and four combinations of $N \in \{256, 1024\}$ and $D \in \{4, 16\}$.
The steepest transition in $p_c(d)$ happens when $\gamma$ is small while $\beta$ is large (cf. lower right panel $\gamma=0.7, \beta=0.5$ in Fig.~\ref{fig:ESN:accuracy:tanh:gamma:beta}).
This is expected since in the previous experiments in
Figs.~\ref{fig:ESN:accuracy:gamma} and~\ref{fig:ESN:accuracy:tanh:beta}, we have seen that both parameters play a role in controlling the recency effect.

\begin{figure*}[t]
\centering
\includegraphics[width=1.0\linewidth]{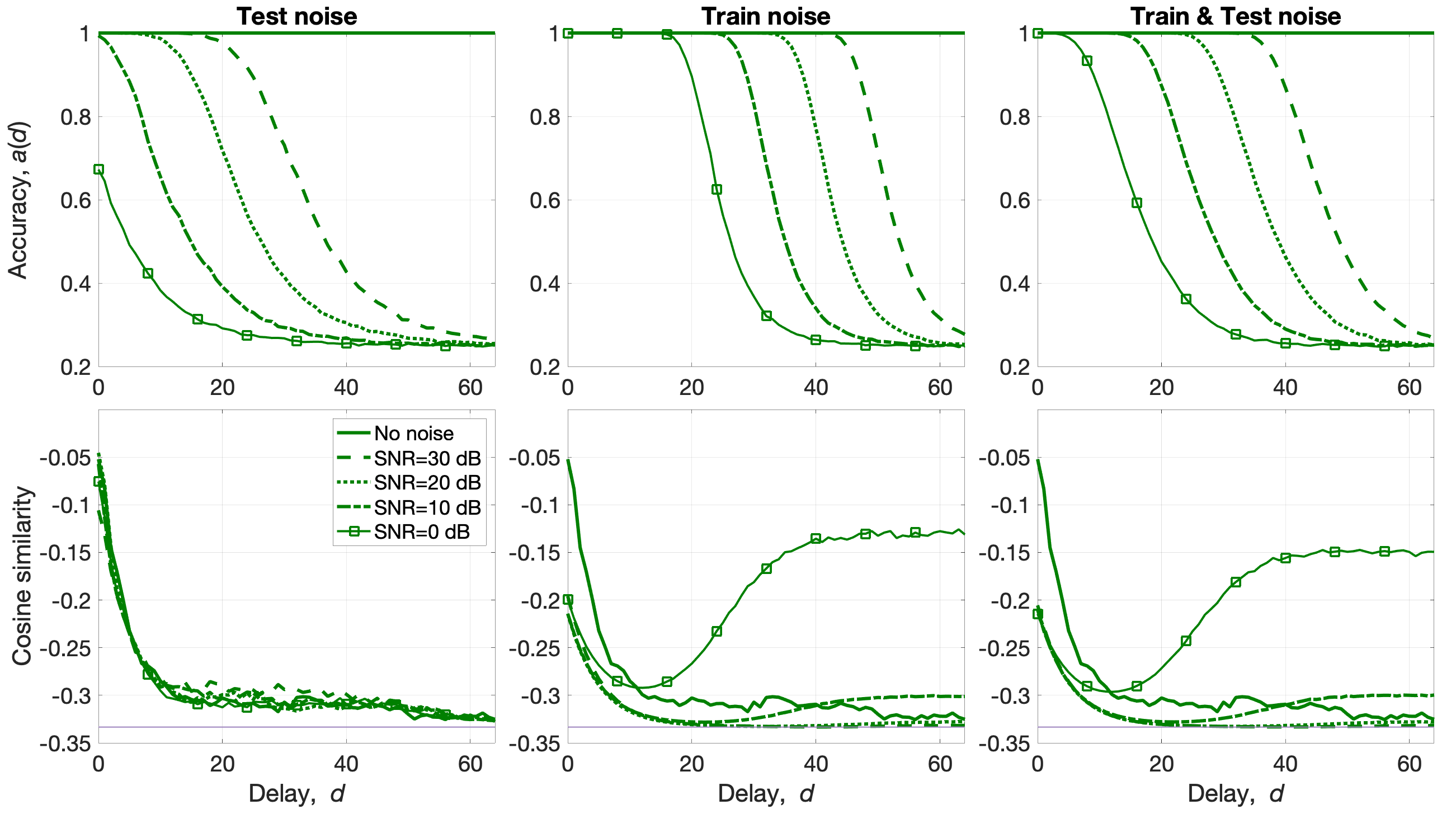}
\caption{
Recall accuracy and cosine similarity between the entries of the regression-based readout matrices;
$N$ was set to $256$; $D$ was set to $4$.
Left column: noise was only added to the reservoir states during the recall phase;
Center column: noise was only added to the reservoir states during the training phase;
Right column: noise was only added to the reservoir states during both test and training phases.
The results were averaged over $20$ simulation runs, each simulation used a random sequence with $E=1,000$, $M=5,000$, and $R=5,000$.
}
\label{fig:ESN:reg:noise}
\end{figure*}

\section{Geometric structure of the readout matrices with noise added to  reservoir states}
\label{readout:reg:noise}

In Fig.~\ref{fig:ESN:readouts:simplex} in the main text, we saw that the for the regression-based readout matrices, the simplex ETF structure was not present neither for small (when $D$ was small) nor for large values of $d$ (when $D$ was large).
We hypothesized that this behavior has to do with the strength of symbols' traces in the reservoir state.
To test this hypothesis we performed the experiment by adding  different amounts of white noise to reservoir states (SNR varied between $\{0,10,20,30\}$ dB). Note that the evolution of the reservoir states was still noiseless.
The noise was added in three natural settings:  only to states during the recall phase, only to states during the training phase, and during both phases. 
Columns in Fig.~\ref{fig:ESN:reg:noise} correspond to these cases. 
As a by-product from the experiment, we can assess the robustness to noise of the regression-based readout matrices in each case. 
So the top row in the figure depicts the recall accuracies while the bottom row presents the cosine similarity between the entries of the readout matrices.

First, when the noise was only added at the recall phase (bottom left panel in Fig.~\ref{fig:ESN:reg:noise}), the cosine similarity between the entries of readout matrix for different amount of noise are overlapping since all readout matrices in these case were obtained from the noiseless reservoir states. 
These curves correspond to the dashed curve in Fig.~\ref{fig:ESN:readouts:simplex} (rightmost panel in the second row).
A different picture is observed once the readout matrices were obtained from nosy reservoir states (bottom center and right panels in Fig.~\ref{fig:ESN:reg:noise}).
Curves for different amount of noise are different from each other. 
Compared to the noiseless case (thick solid line), for small values of $d$, the cosine similarity between the entries of readout matrix obtained from noisy reservoir states was much closer to the simplex ETF structure (thin solid line).
For small amount of noise (SNR=$30$ dB, dashed line), the curve quickly converged to the simplex ETF and stayed there. 
For larger amount of noise (SNR=$\{20, 10\}$ dB), the curves also converged to the simplex ETF quickly but for larger values of $d$ they diverged from it. 
When SNR was $0$ dB, the curve did not even converged to the simplex ETF. 
Thus, the more noise was added to the reservoir states, the more pronounced was the transition to the readout matrix structure different from the simplex ETF.
It is also worth noting that this transition matched the transition in the recall accuracy curves from the high fidelity mode to the random guess.

With respect to the recall accuracy, when the noise was only added at the training phase (top center panel in Fig.~\ref{fig:ESN:reg:noise}), the readout matrix was not able to achieve perfect recall for larger values of $d$.
Moreover, for decreased SNR, the length of the high fidelity mode was decreasing too. 
When the noise was added only at the recall phase (top left panel in Fig.~\ref{fig:ESN:reg:noise}), the recall accuracy was much worse than the noiseless case even when SNR=$30$ dB. 
For example, for SNR=$0$ dB (solid curve with markers), even for $d=0$ the recall accuracy was much lower than one. 
Training the readout matrix on the reservoir states with the same amount of noise as during the recall phase (top right panel in Fig.~\ref{fig:ESN:reg:noise}), improved the situation quite a bit so that even for SNR=$0$ the first several delays had the perfect recall accuracy. 
So, perhaps not surprisingly, the conclusion is that if it is expected that in the recall phase the reservoir states are going to be noisy, the regression-based readout matrix is better to be also obtained from the noisy reservoir states that provide additional regularization~\cite{bishop1995training}.

\section{Assessing geometric structure of the readout matrices with $Z$-scores} 
\label{sect:suppl:zscore}


Section~\ref{sec:results:readouts:structure} in the main text provides some qualitative observations about the geometric structure of the readout matrices. 
Here, we investigate a possible way to aggregate compromises observed for different readouts in a single metric to obtain quantitative observations. 
The idea is to use a $Z$-score, calculated as: 
\noindent
\begin{equation}
Z= \frac{\mu_{h}-\mu_{r}}{\sqrt{\sigma_{h}^2+\sigma_{r}^2}}.
 \label{eq:zscore}
\end{equation}
\noindent
An important assumption, however, is that the distributions of inner products $\mathcal{N}(\mu_{h},\sigma_{h}^2)$ and $\mathcal{N}(\mu_{r},\sigma_{r}^2)$ are considered to be independent.
As indicated by the cosine similarity in the second row in Fig.~\ref{fig:ESN:readouts:simplex}, this is not the case for small values of $D$ but as $D$ increases the independence assumption gets more and more realistic. 
We chose $D=16$ as the cosine similarities between the entries of the readout matrices are close to $0$, but there is still a notable difference between the recall accuracies for different types of readout matrices.

\begin{figure}[t]
\centering
\includegraphics[width=1.0\linewidth]{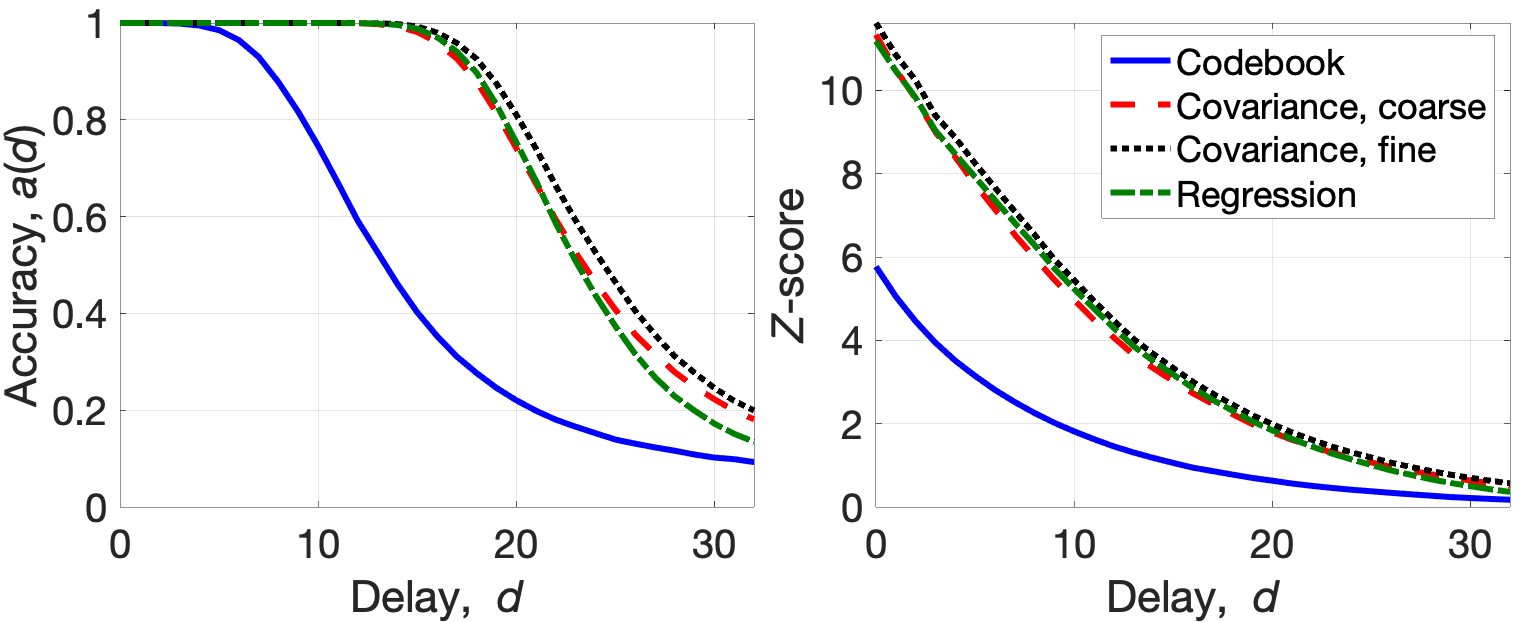}
\caption{
Average accuracy (left panel) and $Z$-scores (right panel) for different types of readout matrices against the delay;
$N$ was set to $256$; $D$ was set to $16$.
The results were averaged over $20$ simulation runs, each simulation used a random sequence with $E=1,000$, $M=5,000$, and $R=5,000$.
}
\label{fig:ESN:readouts:zscore}
\end{figure}

Fig.~\ref{fig:ESN:readouts:zscore} depicts the accuracies (left panel) and the corresponding $Z$-scores (right panel) for $D=16$.
For each readout matrix, the value of $Z$-score was decreasing with increased value of $d$; the decrease in the recall accuracy was following that of $Z$-score once $Z$-score was not high enough to provide the perfect recall. 
As $Z$-scores were approaching $0$, the recall accuracies were approaching a random guess accuracy in ($1/D$). 
The codebook-based readout matrix had the lowest values of $Z$-scores and as a consequence the lowest recall accuracies.  
$Z$-scores of other readout matrices were close to each other but the values for the fine-grained covariance-based readout matrix were slightly higher so its recall accuracy was slightly better too. 
Finally, for small values of $d$, $Z$-scores of the regression-based readout matrix (dash-dotted line) were in between the scores of two covariance-based  types (dashed and dotted lines) so was the recall accuracy.
But once $Z$-scores became smaller than that of the coarse covariance-based readout matrix ($d=23$), the accuracy of the regression-based readout matrix also became worse than that of the covariance-based types.
Thus, $Z$-scores seem to be a viable tool for quantitatively assessing the differences stemming from alternative ways of obtaining readout matrices.

\section{Prediction for echo state networks with the codebook-based readout matrix: 5-Bit Memory Task}
\label{sect:perf:5:bit}

As mentioned in Section~\ref{sect:methods:trajectory}, when slightly modified, $n$-Bit Memory Task corresponds well to the trajectory association task. 
In order to retrieve the cue symbol and the information symbols after $T$ distractor symbols, we should set $d=T$ when forming the readout matrix $\mathbf{W}^{\mathrm{out}}(d)$.
In order to perform well on the task, the network should have high recall accuracy for $d=T$.
In particular, in the experiments in this section, we used $5$-Bit Memory Task and the task was considered successful if the average the accuracy of the network for retrieving all possible $32$ sequences was at least $0.99$.

\begin{figure}[b]
\centering
\includegraphics[width=1.0\linewidth]{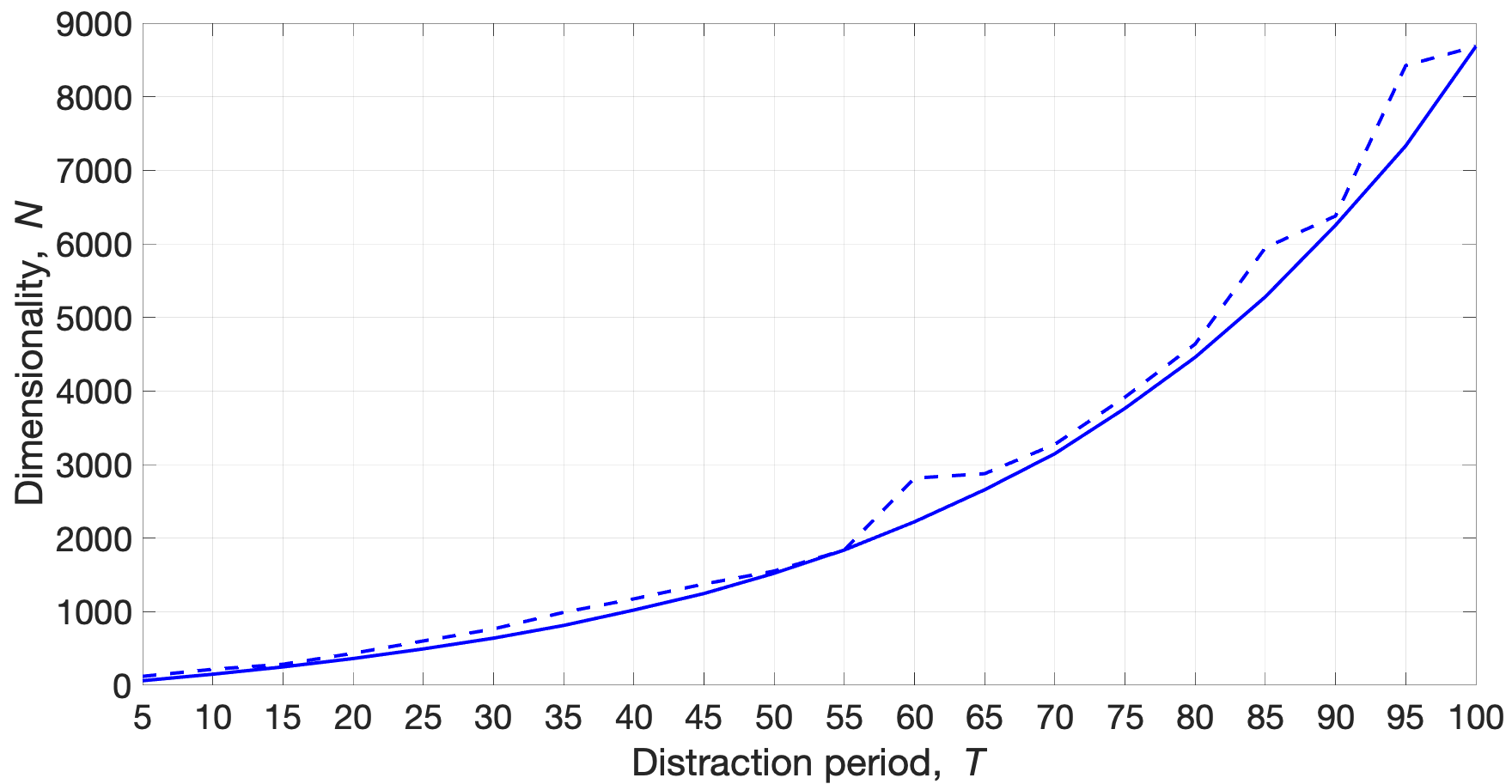}
\caption{
The dimensionality of the reservoir required to succeed at $5$-Bit Memory Task against the size of the distractor period, $T$, which varied in the range $[5, 100]$ with step of size $5$.
The reservoir was updated according to Eq.~(\ref{eq:esnres:var4}); $\mathbf{W}$ was a random permutation matrix; $\mathbf{W}^{\mathrm{out}}(d)$ was the codebook-based readout matrix;
$D=4$ (two information symbols, one distractor symbol, and one cue symbol); $\beta=2^{-6}$; $\gamma=0.99$.
The solid line corresponds to analytical results. 
The dashed line corresponds to empirical results. 
The empirical results were averaged over $50$ simulation runs. 
}
\label{fig:ESN:accuracy:nbit}
\end{figure}

Since the perceptron theory allows us predicting the accuracy of the network for a given value of $d$, it can be used to find such dimensionality of the reservoir, which results in the necessary accuracy for the given length of the distractor period $T$. 
Fig.~\ref{fig:ESN:accuracy:nbit} presents the results for the reservoir updated according to Eq.~(\ref{eq:esnres:var4}) for varying length of the distraction period. 
The figure presents both the analytically predicted (solid line) as well as empirically measured dimensionalities of the reservoir necessary to succeed at $5$-Bit Memory Task.
We see that the required value of $N$ was increasing exponentially with the increased $T$. 
The predicted dimensionalities acted as a lower since bound since the empirically measured dimensionalities were usually slightly larger. 
What is more important, however, is that both curves followed each other very closely and the predictions were able to properly predict the required range of the reservoir size.  

\end{document}